\documentclass[conference]{IEEEtran}
\usepackage{times}

\usepackage[numbers]{natbib}
\usepackage{multicol}
\usepackage[bookmarks=true]{hyperref}
\usepackage{amsmath}
\usepackage{amssymb}
\usepackage{graphicx}
\usepackage{bm}
\usepackage{caption}
\usepackage{lipsum}
\usepackage{booktabs}   
\usepackage{multirow}   
\usepackage{hyperref}
\usepackage{xcolor}

\definecolor{linkblue}{RGB}{0,102,204}

\hypersetup{
    colorlinks=true,
    urlcolor=linkblue
}
\let\labelindent\relax  
\usepackage{enumitem}

\usepackage{booktabs}
\usepackage{makecell}

\newcommand{\methodname}{{\textit{SimToolReal}}}

\begin{document}

\title{\textit{SimToolReal}: An Object-Centric Policy for Zero-Shot
Dexterous Tool Manipulation}

\author{
    Kushal Kedia$^{*1}$ \quad
    Tyler Ga Wei Lum$^{*2}$ \quad
    Jeannette Bohg$^{\dagger 2}$ \quad
    C.~Karen Liu$^{\dagger 2}$ \\[0.6em]
    $^1$Cornell University \quad
    $^2$Stanford University \quad
    $^{*}$Equal contribution \quad
    $^{\dagger}$Equal advising \\[0.8em]
    \href{https://simtoolreal.github.io}
    {\textbf{\textcolor{linkblue}{simtoolreal.github.io}}}
}

\twocolumn[{%
    \renewcommand\twocolumn[1][]{#1}%
    \maketitle
    \vspace{-7mm}
    \begin{center}
      \captionsetup{type=figure}
      \includegraphics[width=\textwidth]{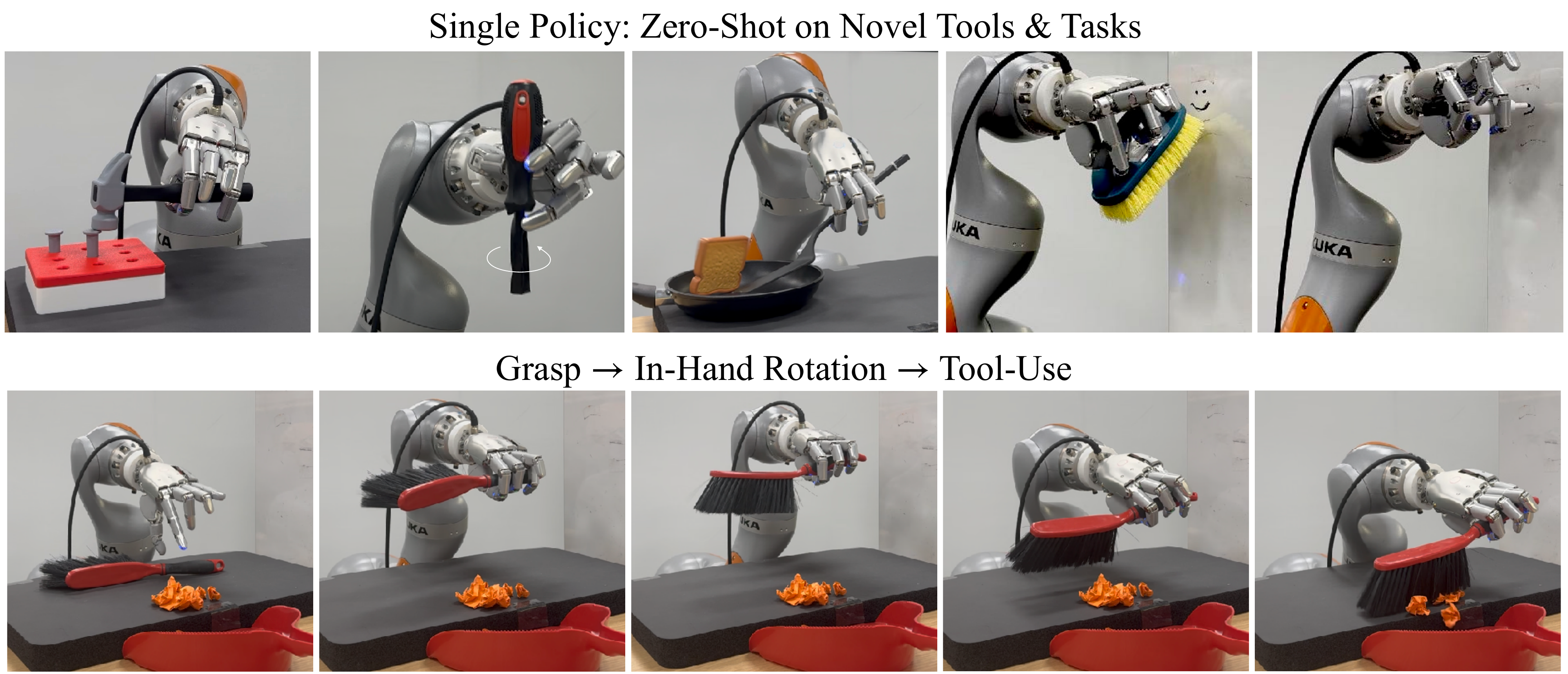}
      \captionof{figure}{
        \textbf{\methodname} is a framework for training a single
        general-purpose, object-centric RL policy in simulation and
        transferring it to real-world tool use. (Top) Zero-shot
        deployment on novel real tools and tasks, spanning thin
        markers to thick hammers. (Bottom) Tool use typically
        involves grasping objects from flat surfaces, reorienting
        them in-hand, and performing the task.
      }
      \label{fig:introduction}
    \end{center}
}]

\thispagestyle{empty}
\pagestyle{empty}

\begin{abstract}
  The ability to manipulate tools significantly expands the set of
  tasks a robot can perform. Yet, tool manipulation represents a
  challenging class of dexterity, requiring grasping thin objects,
  in-hand object rotations, and forceful interactions. Since
  collecting teleoperation data for these behaviors is challenging,
  sim-to-real reinforcement learning (RL) is a promising alternative.
  However, prior approaches typically require substantial engineering
  effort to model objects and tune reward functions for each task. In
  this work, we propose \methodname, taking a step towards
  generalizing sim-to-real RL policies for tool manipulation. Instead
  of focusing on a single object and task, we procedurally generate a
  large variety of tool-like object primitives in simulation and
  train a single RL policy with the universal goal of manipulating
  each object to random goal poses. This approach enables
  \methodname\; to perform general dexterous tool manipulation at
  test-time without any object or task-specific training. We
  demonstrate that \methodname\; outperforms prior retargeting and
  fixed-grasp methods by 37\% while matching the performance of
  specialist RL policies trained on specific target objects and
  tasks. Finally, we show that \methodname\ generalizes across a
  diverse set of everyday tools, achieving strong zero-shot
  performance over 120 real-world rollouts spanning 24 tasks, 12
  object instances, and 6 tool categories. Website: \href{https://simtoolreal.github.io}{simtoolreal.github.io}.
\end{abstract}

\IEEEpeerreviewmaketitle

\section{Introduction}
\begin{figure*}[t!]
  \centering
  \includegraphics[width=0.92\textwidth]{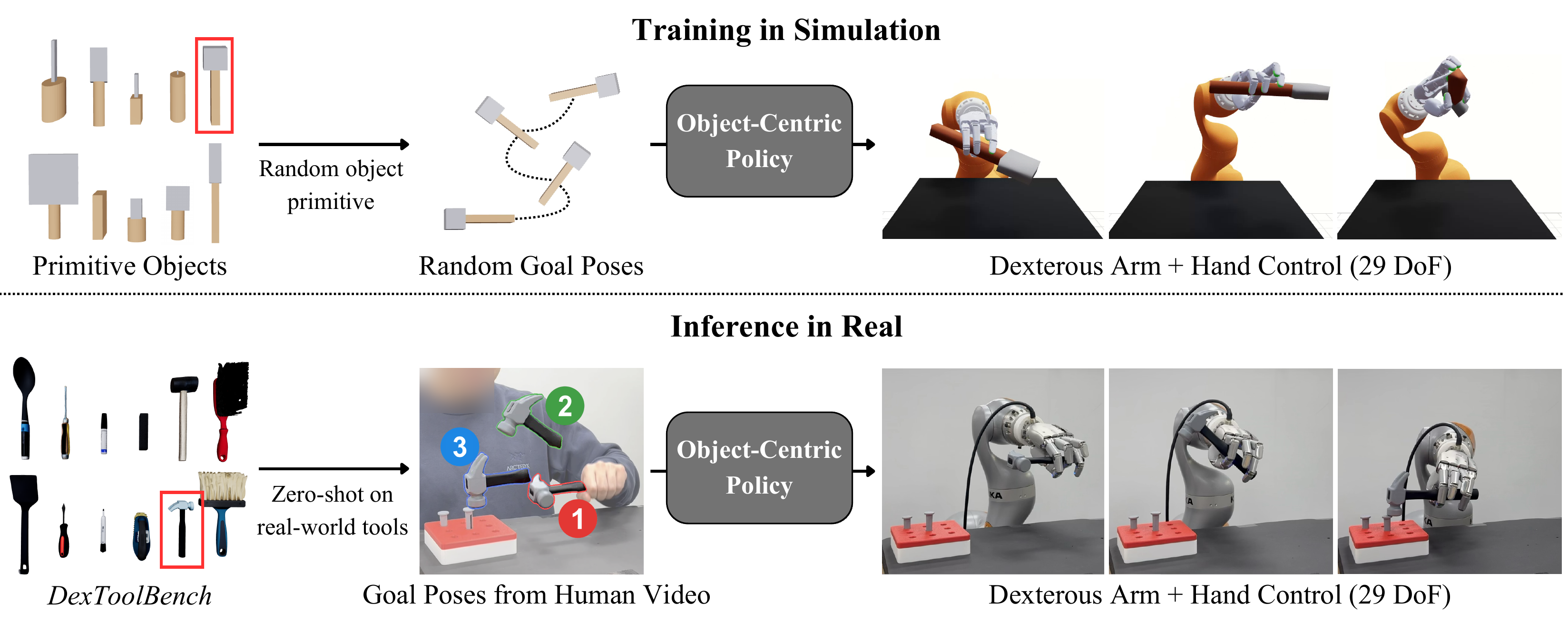}
  \captionsetup{width=\textwidth}
  \caption{\textbf{Overview of \methodname.} (Top) Training in
    Simulation: We train a goal-conditioned RL policy in simulation
    that manipulates a wide variety of procedurally-generated objects
    to randomly sampled goal poses. (Bottom) Inference in Real: We
    deploy this policy zero-shot on real-world tools from
  \textit{DexToolBench}, following tool trajectories from human videos.}
  \label{fig:pipeline}
\end{figure*}

From household chores to industrial work, tool use expands a robot’s
capacity to act on its environment. However, tool use represents a
particularly challenging class of dexterous manipulation tasks.
Success requires grasping tools lying flat on a surface, reorienting
them into functional poses, and maintaining control during
potentially forceful interactions with the environment (see
Fig.~\ref{fig:introduction}). Consider hammering a nail: the robot
must grasp the hammer by its thin handle, rotate it in-hand into a
striking configuration, and deliver impact without losing the grasp.
Such interactions expose a fundamental limitation of parallel-jaw
grippers: with only two opposing contacts along a single grasp axis,
they provide limited resistance against externally induced torques,
making grasp stability primarily rely on friction and grip force.
This motivates multi-fingered dexterous hands for stable tool use.

Imitation learning from teleoperated demonstrations is a common
paradigm for learning manipulation policies~\cite{chi2024diffusion,
zhao2023learning}, but teleoperation is a poor fit for collecting
high-quality dexterous tool-use
data~\cite{chen2025dexforceextractingforceinformedactions}. Human and
robot hands differ in kinematics and actuation, creating a
human-to-robot correspondence
gap~\cite{Si-RSS-24,Human2RobotWholeBodyTransfer} that makes precise
dexterous control unintuitive for the operator~\cite{Gello}.  At the
same time, operators receive limited to no force and tactile
feedback, preventing reliable regulation of contacts during the
task~\cite{okamura2009haptic,
pacchierotti2023cutaneous,HapticFeedbackForSkillLearning,HapticFeedbackSurgery}.

Sim-to-real reinforcement learning (RL) is a promising alternative
that has successfully demonstrated agile, dynamic, and dexterous
behaviors in the real world~\cite{, akkaya2019solving, 11127224}.
Although effective for specific skills, extending its success to
diverse tools and tasks is bottlenecked by substantial per-object
simulation setup and task-specific reward
engineering~\cite{park2024position}. Consequently, existing
approaches are limited to sub-problems of dexterous tool-use, such as
grasping~\cite{lum2024dextrahg,
agarwal2023dexterousfunctionalgrasping, chen2025clutterdexgrasp},
in-hand object reorientation~\cite{Chen_2023,
openai2019learningdexterousinhandmanipulation,handa2023dextreme}, or
spinning objects~\cite{wang2024penspin, lin2024twisting}.

Our goal is to learn a dexterous manipulation policy that can be
applied zero-shot to novel tools and tasks at test-time. Our key
insight is to view dexterous tool use from an \textit{object-centric}
lens, framing a task as moving a tool through a sequence of goal
poses. Thus, we focus on training a single goal-conditioned policy
that manipulates diverse objects to random goal poses. In simulation,
we procedurally generate a large set of primitive objects and train
one unified goal-conditioned policy. At test time, we compose this
policy for zero-shot tool use by sequentially conditioning it on a
goal pose trajectory extracted from a human video~\cite{dan2025x,
lum2025crossinghumanrobotembodimentgap}. This provides dense guidance
from the initial grasp, through in-hand reorientation into a
functional configuration, and into the ensuing tool-use motion. Our
approach avoids per-task reward shaping while allowing generalization
to novel tools and trajectories at test-time. Intuitively, mastering
the ability of manipulating objects to any random pose induces the
core skills required for tool use: establishing an initial grasp,
in-hand object reorientation, and maintaining stable contact.

We instantiate this approach in \methodname, a framework for training
a single general-purpose RL policy in simulation and deploying it for
real-world tool use (Fig.~\ref{fig:pipeline}). To enable
generalization to novel objects, the policy is conditioned on the
tool’s current 6D pose and a coarse 3D bounding box over its
graspable region (e.g., the handle of a hammer). This abstraction
enables zero-shot transfer to the real-world by effectively bypassing
the sim-to-real visual gap. At deployment, we recover this
representation on real tools using a perception pipeline based on
vision foundation models, combining SAM 3D~\cite{chen2025sam} for
segmentation and mesh extraction with
FoundationPose~\cite{wen2024foundationpose} for 6D pose tracking.

To evaluate generalization, we introduce \textit{DexToolBench}, a
benchmark of daily tool-use behaviors, both in simulation and in the
real-world. Each task is paired with a human video demonstration, and
the robot is evaluated on its ability to follow the demonstrated tool
trajectory. Notably, \methodname\; is trained exclusively on
procedurally-generated primitives with randomly sampled goal pose
trajectories. Yet, it transfers zero-shot to the trajectory following
tasks in \textit{DexToolBench} and matches or outperforms specialist
policies trained on a single object instance and task. Our contributions:
\begin{enumerate}[leftmargin=*]
  \item We propose \methodname{}, a unified RL framework that frames
    dexterous tool use as manipulating a tool through a sequence of
    desired poses. Training a single policy under this objective
    induces dexterous skills, such as stable grasping and in-hand
    object reorientation. This policy is used zero-shot for tool-use
    without task-specific engineering.
  \item We develop an object-centric perception pipeline that
    estimates the tool’s 6D pose and a coarse 3D grasp bounding box
    using vision foundation models, enabling zero-shot sim-to-real
    deployment of the RL policy.
  \item We introduce \textit{DexToolBench}, a benchmark of dexterous
    tool-use tasks. We show that \methodname\ achieves strong
    zero-shot performance on this benchmark over 120 real-world
    rollouts spanning 24 tasks, 12 object instances, and 6 tool
    categories. We outperform prior methods using fixed grasps and
    motion retargeting by 37\% in task progress while matching the
    performance of specialist policies.
\end{enumerate}

\section{Related Work}
\textbf{Imitation Learning from Teleoperated Data.} Imitation
learning (IL) is the predominant approach to train robot manipulation
policies~\cite{ze20243ddiffusionpolicygeneralizable,
lin2024learning,cheng2025moe}. However, IL relies on high-quality
demonstrations, typically collected via human teleoperation. Many
systems have been proposed to teleoperate multi-fingered dexterous
hands via motion capture gloves~\cite{shaw2024bimanual,
wang2024dexcap}, VR devices~\cite{bunny-visionpro,
iyer2024open,lin2025learning, arunachalam2022holo}, or even direct
camera input~\cite{qin2024anyteleopgeneralvisionbaseddexterous,
handa2019dexpilotvisionbasedteleoperation, sivakumar2022robotic}.
However, directly mapping hand motion to robot actions can lead to
imprecise, unintuitive
control~\cite{chen2025dexforceextractingforceinformedactions,
Si-RSS-24,Human2RobotWholeBodyTransfer}. Further, operators receive
limited haptic and tactile feedback during teleoperation, which
limits the collected demonstrations to simple manipulation
tasks~\cite{okamura2009haptic,
pacchierotti2023cutaneous,HapticFeedbackForSkillLearning,HapticFeedbackSurgery}.

Recent works propose wearable exoskeleton systems that physically
couple the robot hand to the operator during data
collection~\cite{tao2025dexwild, xu2025dexumi, fang2025dexop}. Still,
these bulky exoskeleton systems carry a considerable burden on the
operator, constrain the range of human motion, and fail to
demonstrate complex skills, such as in-hand reorientations. Another
line of work treats the problem of teleoperation as shared autonomy.
These approaches learn structured action-spaces from human hand
motion data~\cite{naughton2024respilot} or RL in
simulation~\cite{hsieh2025learning, liu2025dexndmclosingrealitygap},
which are used by the human operator to control the robot. Recently,
DexterityGen~\cite{yin2025dexteritygenfoundationcontrollerunprecedented}
trains RL policies conditioned on target grasp poses across diverse
objects. However, unlike our approach, it trains object-specific
policies and distills them into a model that only conditions on
hand-motion inputs. Largely, shared-autonomy methods are
\textit{object-blind} at execution time and rely on skilled operators
to bridge the gap between perception and control. In contrast to all
these approaches, we focus on autonomous manipulation policies
without any human teleoperation effort.

\textbf{Learning Dexterity from Human Video Demonstrations.} Learning
dexterous manipulation directly from human video demonstrations is a
promising alternative to teleoperation. Kinematic motion retargeting
pipelines can translate human hand motions into robot actions by
reconstructing hand–object interaction~\cite{okami2024,
lakshmipathy2025kinematic}. More recently, these pipelines have been
extended to convert demonstrations synthesized by video-generation
models into robot-executable trajectories~\cite{patel2025robotic,
li2025novaflow, chen2025large}. However, because these methods
primarily rely on kinematic hand-motion references, they struggle to
produce actions that reliably grasp and maintain contact with tools.

To address this limitation, another line of work performs
\textit{functional retargeting}, focusing on matching the
demonstrated task outcome rather than precisely matching human hand
motion. For instance, hand–object 3D motion-capture can be used to
train dexterous RL policies in simulation that reproduce the object
motion while discovering robot-specific contact
strategies~\cite{mandi2025dexmachinafunctionalretargetingbimanual,
chenobject, li2025maniptrans}. More recently, \textit{functional
retargeting} has been extended to RGB-D human video
demonstrations~\cite{lum2025crossinghumanrobotembodimentgap,
dan2025x} and even AI-generated videos~\cite{mao2025robot}. Despite
their success, these approaches typically require time-intensive RL
training where the resulting policy is limited to the demonstrated
manipulation task and object. In contrast, our approach can imitate
object motion directly from a single RGB-D human demonstration
without any additional demonstration-specific training. We can use
the same policy across diverse tasks.

\textbf{Sim-to-Real Learning.} Policies learned in simulation can
acquire dexterous manipulation behaviors without relying on human
demonstrations, but transferring these skills to the real world often
requires substantial \textit{environment
shaping}~\cite{park2024position}. This includes closing the
sim-to-real gap, building accurate task-specific object models, and
carefully designing and tuning rewards for each behavior. Domain
randomization over physical and visual
parameters~\cite{handa2023dextreme, akkaya2019solving,
andrychowicz2020learning} and adaptation methods that infer
real-world dynamics online~\cite{kumar2021rma, qi2022hand,
uppal2024spin} can reduce the sim-to-real gap significantly. Still,
sim-to-real dexterous manipulation requires per-task engineering,
which has led many systems to focus on narrower skill subsets such as
grasping~\cite{agarwal2023dexterousfunctionalgrasping, ye2025dex1b,
lum2024dextrahg, singh2025end, singh2024dextrah}, in-hand object
reorientation~\cite{Chen_2023, liu2025dexndmclosingrealitygap}, or
dynamic object spinning~\cite{wang2024penspin, lin2024twisting}.
Recent research in locomotion has begun to unify RL policies across
embodiments by leveraging large-scale simulation with procedurally
generated robots and unified reward
functions~\cite{liu2025locoformer, ai2025towards}. Inspired by this
direction, we train a single RL policy across procedurally generated
object primitives and the universal reward function of reaching
random goal poses. We show that this formulation induces highly
dexterous skills (e.g., in-hand reorientation and dynamic spinning)
and transfers to diverse real-world tools without task-specific
engineering effort.

\begin{figure*}[t!]
  \centering
  \includegraphics[width=\textwidth]{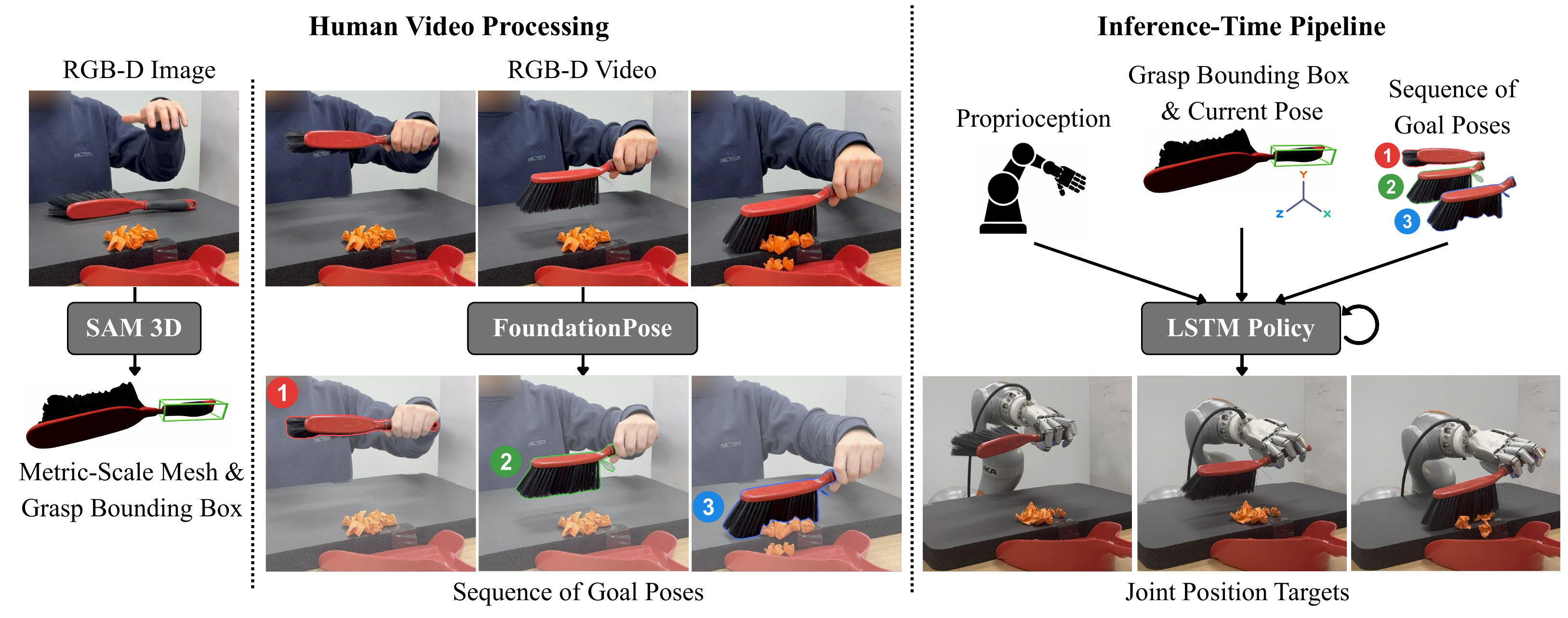}
  \captionsetup{width=\textwidth}
  \caption{\textbf{Real-World Deployment.} (Left) Human Video
    Processing: We collect an RGB-D human video and process it using
    vision foundation models. We use SAM 3D~\cite{chen2025sam} to
    generate a metric-scale object mesh and segment a 3D grasp
    bounding box. Then, we use
    FoundationPose~\cite{wen2024foundationpose} to extract a sequence
    of 6D goal poses. (Right) Inference-Time Pipeline: Our LSTM
    policy takes in proprioception, object pose, grasp bounding box,
    and goal pose, and it outputs joint position targets for the
    29-DoF dexterous robot (arm + hand).
  }
  \label{fig:inference}
\end{figure*}

\section{SimToolReal}

We propose \methodname, a sim-to-real RL framework for zero-shot,
dexterous tool manipulation that generalizes to unseen tools and
manipulation sequences. Our key insight is to view dexterous tool-use
tasks through an \textit{object-centric lens}: a wide range of tasks
can be specified as manipulating a tool through a sequence of goal
poses. This reduces the tool-use problem to learning a single
goal-reaching RL policy in simulation that manipulates procedurally
generated tool primitives toward random goal poses. This training
objective induces dexterous behaviors, such as stable grasping and
in-hand reorientation, that are essential for tool use. At test time,
the policy takes as input a sequence of object goal poses (e.g., from
a human video) and executes actions to track this sequence, enabling
diverse zero-shot tool-use without any real-world object modeling in
simulation or task-specific training.

\subsection{Problem Formulation}
We consider the problem of training dexterous policies that can
manipulate any tool to arbitrary goal poses. Let $\bm{o}_t \in SE(3)$
denote the current object pose at time $t$, $\bm{s}_t$ denote robot
proprioception, and $\bm{\phi}$ denote a coarse object descriptor
(e.g., approximate geometry or physical properties). Given a goal
pose $\bm{g} \in SE(3)$, we learn a policy $\bm{a}_t =
\pi_\theta\!\left(\bm{s}_t,\, \bm{o}_t,\, \bm{\phi},\, \bm{g}\right)$
that outputs joint position targets for the robot's arm and hand.
Task execution then reduces to repeatedly reaching the current goal
pose and advancing to the next goal. This abstraction lets a single
goal-reaching controller support diverse tool-use tasks, without
task-specific policy training or reward design.

To perform a tool-use task, we assume access to a sequence of goal object poses
$\{\bm{g}^k\}_{k=1}^{K}$ with $\bm{g}^k \in SE(3)$, extracted from a
human demonstration video. For example, in Fig.~\ref{fig:pipeline},
we visualize the goal sequence of reorienting the hammer, aligning it
above the nail, and swinging. During execution, we condition the
policy on the current goal pose $\bm{g}^k$ and advance to the next
goal $\bm{g}^{k+1}$ when the object pose $\bm{o}_t$
is sufficiently close, i.e., when $d(\bm{o}_t,\bm{g}^k) < \epsilon$.
Here, $d$ is a pose distance function and $\epsilon$ is the success threshold.

In the following, we describe how we train our goal-conditioned RL
policy in simulation, and the sim-to-real pipeline used for
real-world deployment.

\subsection{Training a General-Purpose Goal-Reaching Policy}
\label{sec:training}

\textbf{Environment Setup.} We design a general environment to model
tool-use tasks. At the start of each episode, we place a randomly
selected object on the table in a random pose and initialize the
robot in a randomized joint configuration. We then sample a sequence
of goal object poses
$\{\bm{g}^k\}_{k=1}^{K}$. The robot must first learn to grasp the
object from the flat table surface and subsequently manipulate it to
reach each goal. If the object is dropped, the episode terminates and
is reset. We sample the first goal randomly in the robot's reachable
workspace to expose the robot to a wide range of object poses and
large reorientations. Subsequent goals are sampled close to the
previous goal to encourage smooth, trajectory-like motion.

\textbf{Reward Function.} Our primary objective is to train a
goal-pose reaching RL policy. Practically, we
optimize~\cite{petrenko2023dexpbt}:
\begin{equation}
  r = r_{\textrm{smooth}} + r_{\textrm{grasp}} +
  \mathbb{I}_{\textrm{grasped}}r_{\textrm{goal}}.
\end{equation}
$r_{\textrm{smooth}}$ is a regularization term that encourages smooth
actions, $r_{\textrm{grasp}}$ is a shaping term that encourages
grasping the object at the start of the episode, and
$\mathbb{I}_{\textrm{grasped}}$ indicates whether the object has been
grasped. After the object is grasped, $r_{\textrm{goal}}$ is the
dominant reward term, driving goal-pose reaching. We define:
\begin{equation}
  r_{\rm goal}
  = \textrm{max}(d^* - d(\bm{o}_t,\bm{g}), 0) + B_{\rm
  succ}\,\mathbb{I}\!\left[d(\bm{o}_t,\bm{g}) < \epsilon\right],
\end{equation}
which consists of (i) a dense term that minimizes the distance
$d(\bm{o}_t, \bm{g})$ to the goal, and (ii) a large, sparse success
bonus $B_{\rm succ}$. $d^*$ is a stateful variable that tracks the
minimum distance achieved so far, which is reset when a new goal is
sampled. Positive reward is only provided for progress towards the
goal, preventing the agent from exploiting the reward function by
maintaining static proximity. Whenever the goal pose is reached
(i.e., $d(\bm{o}_t,\bm{g}) < \epsilon$), the agent receives the
success bonus and a new goal is sampled.
Following~\cite{allshire2022transferring,petrenko2023dexpbt}, we
define the distance to the goal pose as $d(\bm{o}_t,\bm{g}) = \max_{i
}\|\bm{o}_{t, i} - \bm{g}_i\|$, where we represent each pose using
$D=4$ keypoints in its local frame. $\bm{o}_{t, i}$ and $\bm{g}_i$
denote the positions of the $i$-th keypoint of the current and goal
poses, respectively. See Appendix~\ref{app:reward} for more details
about the reward terms.

\textbf{Procedural Generation of Tools in Simulation.} We generate a
large diversity of tool-like primitive objects in simulation such
that they span the variation seen in real-world tools. Each tool
primitive is generated as a combination of a \textit{handle} and a
\textit{head}: we sample their geometries as cylinders and cuboids
with varying dimensions. While simple, this design captures the key
structure of many everyday tools, such as brushes, spatulas, markers,
and hammers. A random subset of our generated primitives is
visualized in Fig~\ref{fig:pipeline}. In addition to geometric
variation, we randomize mass distribution by assigning different
densities to the handle and head, representative of many real-world
tools (e.g., hammer heads are typically much denser than handles).
Additional generation details are provided in Appendix~\ref{app:asset_gen}.

\textbf{Object-Centric Policy Inputs.} To deploy a single policy
across diverse real-world tools, we need an object representation
that makes learning efficient while also being feasible to extract in
the real-world. Real-world tools are diverse in their geometry and
physical properties, and accurately estimating detailed geometric or
physical parameters from visual sensing alone is impractical.
Instead, we provide the policy only the object inputs that are
reliably available at deployment: the current 6D tool pose and a
coarse 3D grasp-region bounding box. The bounding box encodes the
intended graspable region (center + extents in the object frame;
Fig.~\ref{fig:inference}) and is held fixed during an episode. We use
an LSTM backbone so the policy can integrate interaction history and
implicitly infer latent physical and geometric properties that are
not directly observed. This design follows the spirit of prior work
such as DexFunc~\cite{agarwal2023dexterousfunctionalgrasping} and
RMA~\cite{qi2022hand}, which demonstrate dexterous manipulation
behaviors with limited object perception to reduce the sim-to-real
observation gap.

\textbf{RL Training Details.} We highlight key RL design choices
essential for policy learning and sim-to-real transfer.

\noindent\textit{a) SAPG}~\cite{sapg2024}: We train our policy using
SAPG, a variant of PPO~\cite{PPO}. We find that standard PPO
encounters exploration bottlenecks in massively parallel simulations.
SAPG mitigates this by maintaining a population of policies to
promote exploration diversity and updating the leader policy using
their collective experience.

\noindent\textit{b) Domain Randomization}: We apply targeted domain
randomization during training to aid sim-to-real transfer. This
includes observation delays, action-execution latency, noise and
delay in object-pose estimates, and perturbations to the grasp-region
bounding box. We additionally apply random force and torque
perturbations to the object, encouraging the policy to learn strong,
stable grasps. Randomization ranges and other details are provided in
Appendix~\ref{app:sim-to-real}.

\noindent\textit{c) Asymmetric
Critic}~\cite{asymmetric_actor_critic}: While the actor receives only
the minimal object representation available at test time, we train
with an asymmetric actor–critic setup in which the critic has access
to privileged states in simulation. We also provide the critic with
clean (noise-free, undelayed) observations to improve value
estimation and stabilize training. Additional details are included in
Appendix~\ref{app:sim-to-real}.

\begin{figure*}[t!]
  \centering
  \includegraphics[width=\textwidth]{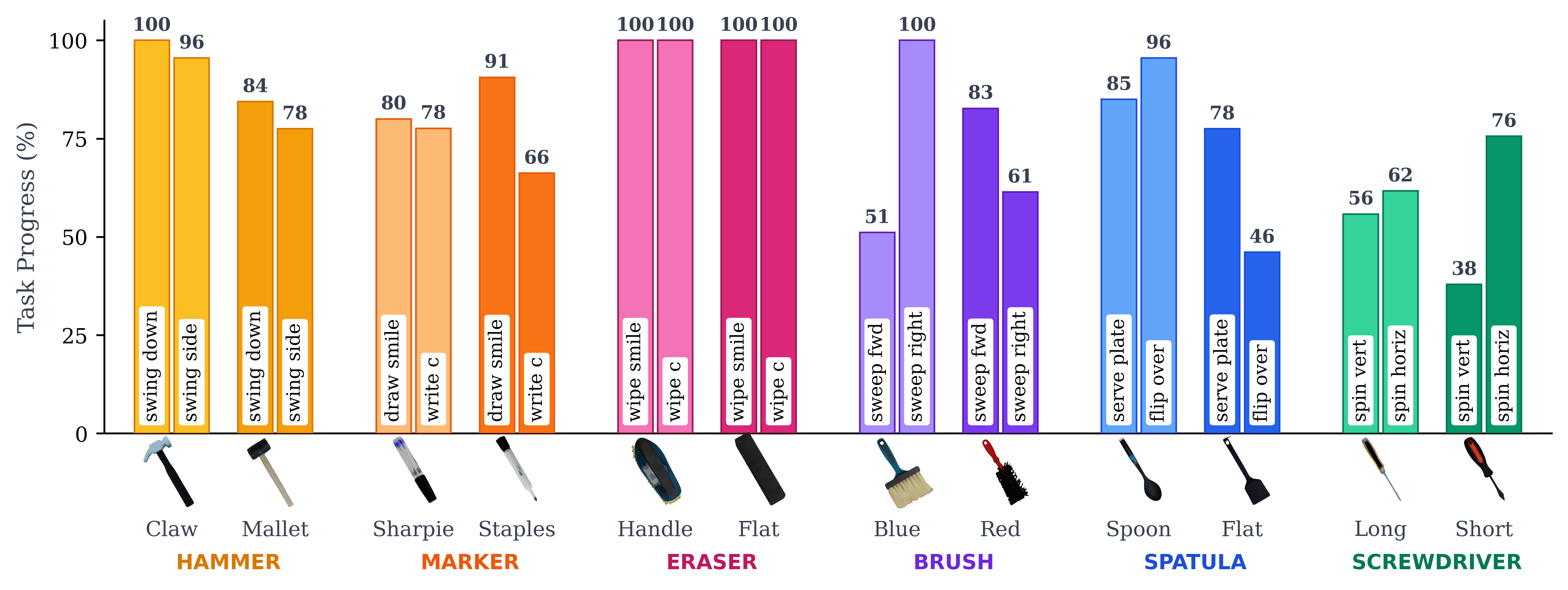}
  \captionsetup{width=\textwidth}
  \caption{\textbf{Generalization to Unseen \textit{DexToolBench}
    Tools and Tasks in the Real World.} We evaluate our policy in the
    real world on unseen tool-use tasks in \textit{DexToolBench}. Our
    evaluations span 24 unique task trajectories across 6 different
    object categories and 12 object instances. Each bar corresponds to
    1 task trajectory on 1 object instance. We report the average
    \textit{Task Progress} across 5 rollouts.  Despite not being
    trained on these objects or trajectories, our policy demonstrates
  strong generalization to diverse tools of varying masses and geometries.}
  \label{fig:generalizationResults}
\end{figure*}

\subsection{Real-World Deployment}
In this section, we describe how we deploy our object-centric policy
in the real world to perform a wide range of tool-use tasks.
Fig.~\ref{fig:inference} provides an overview. Given a third-person
RGB-D view of a human demonstration, we extract (i) a 3D metric-scale
object mesh and grasp region bounding box, and (ii) a sequence of
target object poses that serves as the goal trajectory. During
execution, we run the policy in a closed loop to reach the goal poses
sequentially.

In contrast to~\citet{lum2025crossinghumanrobotembodimentgap}, our
approach does not require access to the real-world object or target
trajectory during training, enabling zero-shot generalization to
novel objects using demonstrations collected entirely after training.

\textbf{Human Video Processing.} From the first RGB-D frame, we
reconstruct a metric-scale 3D mesh of the object using SAM
3D~\cite{chen2025sam}. We then segment the intended grasp region
using SAM 2~\cite{ravi2024sam2}, and convert this region into a
coarse 3D bounding box that is provided as part of the policy input.
Finally, we run FoundationPose~\cite{wen2024foundationpose} on the
RGB-D video conditioned on the extracted mesh to obtain the 6D
object-pose trajectory. We process this trajectory by downsampling to
3 Hz. We use this processed trajectory as the goal sequence at test
time. See Appendix~\ref{app:human_video_processing} for implementation details.

\textbf{Inference-time Object Tracking.} During inference, we
estimate the current 6D object pose with an update frequency of 30
Hz. For live pose tracking, we use
FoundationPose~\cite{wen2024foundationpose} with RGB-D observations
from a third-person camera view combined with the 3D mesh extracted
from the human video. At each control step, the policy is conditioned
on proprioception, the current object pose, the (fixed) grasp-region
bounding box, and the current goal pose from the demonstration
trajectory. We initialize the goal as the first pose in the sequence,
switching it to the next goal whenever the pose error
$d(\bm{o}_t,\bm{g})$ falls within a tolerance $\epsilon$. The episode
ends when the final goal is reached or upon failure (e.g., loss of
grasp or tracking).

\subsection{\textit{DexToolBench}}
\label{section:dex_tool_bench}

Existing robot manipulation benchmarks primarily focus on robots with
parallel-jaw
grippers~\cite{luo2024fmbfunctionalmanipulationbenchmark,jain2025polarisscalablerealtosimevaluations,atreya2025roboarena,pumacay2024colosseum,james2019rlbench,9156992,li2024behavior1k}
or dexterous
grasping~\cite{ye2025dex1b,wang2025dexh2rbenchmarkdynamicdexterous,zhangdexgraspnet,xu2023unidexgrasp,wan2023unidexgrasp++}.
We introduce \textit{DexToolBench}: a real-world, dexterous
manipulation benchmark of challenging tool-use tasks paired with
digital-twin simulation environments. These tasks require grasping
tools from flat surfaces, in-hand rotations to functional
configurations, and environment interactions. This benchmark consists
of 24 daily tool-use tasks, 12 unique object instances across 6 categories.

Each task is defined by an RGB-D human video, captured via a ZED 1
stereo camera. We process each video with our perception pipeline
(Fig.~\ref{fig:inference}) to extract a 3D object mesh and a 6D
object pose trajectory. To facilitate reproducibility, we provide the
raw RGB-D videos, processed data, simulation environments, and
purchase links for the real-world tools.

Each tool instance is visualized in
Fig~\ref{fig:generalizationResults}. We briefly describe the skills
required in each tool-use category in the benchmark:

\noindent \textit{a) Hammer:} Grasp, 90$^\circ$ in-hand rotation,
swing down or side.

\noindent \textit{b) Marker:} Grasp a thin object, write on a whiteboard.

\noindent \textit{c) Eraser:} Grasp, wipe marker ink on whiteboard.

\noindent \textit{d) Brush:} Grasp, 90$^\circ$ in-hand rotation,
sweep forward or right.

\noindent \textit{e) Spatula:} Grasp, serve or 180$^\circ$ in-hand
rotation and flip.

\noindent \textit{f) Screwdriver:} Grasp, 90$^\circ$ in-hand
rotation, spin in free space along vertical or horizontal axis.

See Appendix~\ref{app:dex_tool_bench} for more details about
\textit{DexToolBench}.

\section{Experiments}

\textbf{Experimental Setup}: We evaluate \methodname\; on a robot
consisting of a 22-DoF Sharpa five-fingered left hand mounted on a
7-DoF KUKA iiwa 14 arm. We evaluate our policies on
\textit{DexToolBench} (Sec.\ref{section:dex_tool_bench}). For each
task, we evaluate \textit{Task Progress}, measuring the percentage of
demonstrated goal poses tracked successfully. We focus on trajectory
following of a fixed goal sequence rather than evaluating functional
task completion. We consider a goal pose to be reached if the
distance between the object pose and goal pose $d(\bm{o}_t,\bm{g})$
is below our defined success tolerance $\epsilon=2cm$. We design
experiments to answer the following questions:
\begin{enumerate}[leftmargin=*, label=\Alph*.]
  \item What is the zero-shot transfer performance of \methodname{}
    on unseen objects and trajectories in the real world?
  \item How does \methodname\; compare to prior methods that use
    kinematic motion retargeting or fixed-grasps?
  \item How does \methodname\ compare to task-specific specialists
    trained on a single object and trajectory?
  \item To what extent does our training objective correlate with
    downstream performance on unseen tools and tasks?
  \item Which RL design decisions are important for maximizing
    training performance?
\end{enumerate}

\begin{figure*}[t!]
  \centering
  \includegraphics[width=\textwidth]{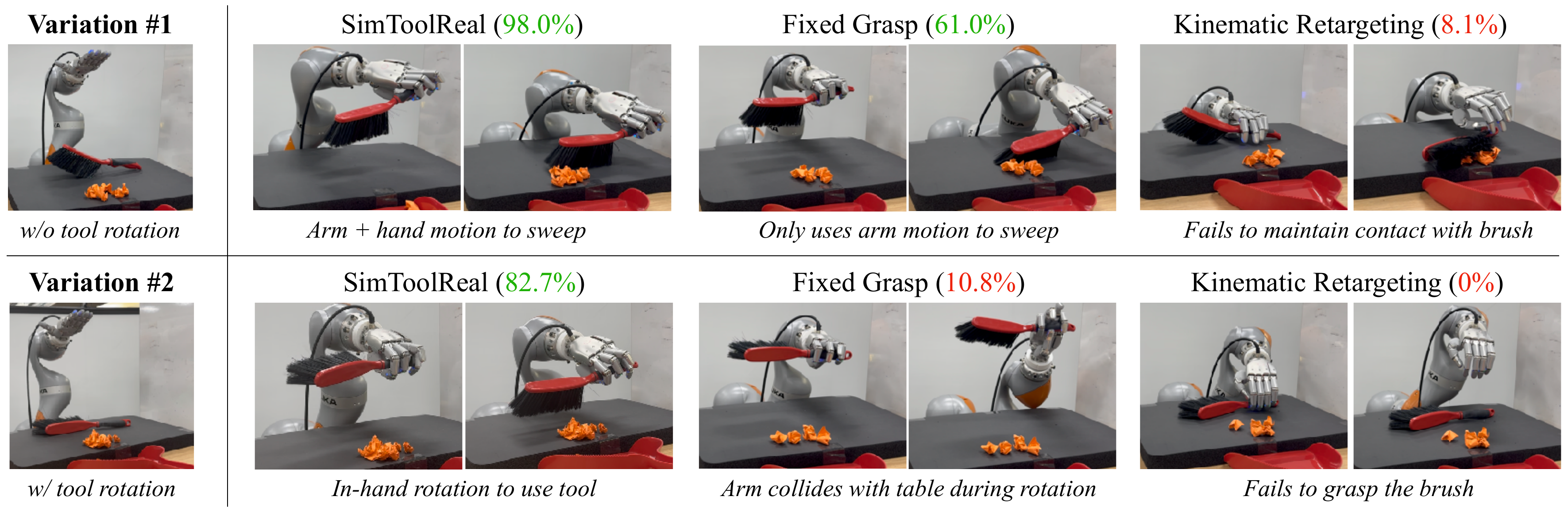}
  \captionsetup{width=\textwidth}
  \caption{\textbf{Comparison against Baselines in the Real World.}
    We compare \methodname\; against baselines on two variations of
    sweeping a table with a brush: with and without requiring tool
    rotation based on the initial states shown on the left. Average
    \textit{Task Progress} is indicated in parentheses. \methodname\;
    succeeds on both variations, performing dexterous in-hand tool
    rotations in the harder variation. \textit{Fixed Grasp} succeeds
    on the simpler variation of this task without tool rotation.
    However, when rotation is required, enforcing a fixed grasp
    causes the arm to collide with the table while tracking the
    target trajectory.
    \textit{Kinematic Retargeting} fails to reason about contact
  forces, and is unable to grasp the brush in both variations.}
  \label{fig:dexFuncComparison}
\end{figure*}

\subsection{Zero-Shot Real-World Tool-Use}

\textbf{Setup.} We evaluate whether our single policy trained in
simulation transfers zero-shot to real-world tool use, without any
object- or task-specific fine-tuning. As shown in
Fig.~\ref{fig:generalizationResults}, our evaluation spans 24 task
trajectories across 6 tool categories and 12 object instances in
\textit{DexToolBench} (Sec.~\ref{section:dex_tool_bench}), all of
which were unseen during training. We report average \textit{Task
Progress} over 5 trials, totaling 120 real-world rollouts.

\textbf{Results.} Overall, the policy demonstrates strong zero-shot
generalization across tools with diverse masses and geometries. We
observe the highest \textit{Task Progress} on \texttt{eraser}
trajectories, which rely primarily on translation motion rather than
in-hand object rotation. Similarly, \texttt{marker} trajectories
don't require in-hand rotations, but because the tool is much
thinner, the grasps are less reliable. Further, the marker's small
size makes it prone to pose tracking loss during occlusion.

The remaining four categories (\texttt{hammer}, \texttt{brush},
\texttt{spatula}, \texttt{screwdriver}) require substantial in-hand
object rotation to reach functional tool configurations. While the
policy still achieves strong \textit{Task Progress} across these
tasks, performance degrades on thinner tools (e.g., performs better
  on the $\sim$3cm thick \texttt{spoon spatula} than the $\sim$1cm
thick \texttt{flat spatula}) and heavier tools (e.g., performs better
  on the 36g \texttt{claw hammer} than the 331g \texttt{mallet
hammer}). We also note that tool geometry impacts task difficulty.
For instance, the \texttt{red brush}'s sideways head makes
\texttt{sweep forward} easier, but the \texttt{blue brush}'s frontal head
makes \texttt{sweep right} easier. The \texttt{screwdriver} is the
most challenging category, as the task requires both functional
reorientation and continuous spinning.

\textit{Failure Analysis.} The most common failure mode was pose
tracking loss (43.7\% of failures), followed by object drops
(34.5\%), failure to reach the goal pose due to incomplete in-hand
rotation (18.2\%), and grasp failure (3.6\%). See
Appendix~\ref{app:real_world_analysis} for additional quantitative
and qualitative analysis of our results. Notably, we observe strong
recovery behavior when the policy makes mistakes. For example, when
the object is dropped, the policy consistently attempts to re-grasp
the object, provided the object remains within the workspace and pose
tracking is not lost.

\subsection{Comparisons to Retargeting and Fixed Grasp Baselines}

\textbf{Setup.} In this section, we compare \methodname\; with:

\noindent \textit{(i) Kinematic Retargeting:} Following prior
works~\cite{okami2024,lum2025crossinghumanrobotembodimentgap,guzey2024bridginghumanrobotdexterity,qiu2025-humanpolicy,
pavlakos2024reconstructing}, we retarget hand motion from a human
video into dexterous robot actions. From an RGB-D video, we estimate
3D hand finger positions and then solve for arm and hand joint
positions using an Inverse Kinematics (IK) solver. Although this
method bypasses the need for RL, the resulting motion is purely
kinematic and fails to account for contact interactions. See
Appendix~\ref{app:kinematic_retargeting} for implementation details.

\noindent \textit{(ii) Fixed Grasp:} Recent
works~\cite{agarwal2023dexterousfunctionalgrasping,patel2025robotic,hsu2025spotse3posetrajectory,zhu2024orion,allu2025hrt1oneshothumantorobottrajectory}
frame manipulation tasks as establishing a grasp, then subsequently
fixing the grasp while moving the object. To perform any object
rotations, the robot must strictly rely on arm motion. This baseline
isolates the necessity of in-hand manipulation; failure here
demonstrates that the arm lacks the kinematic workspace to rotate
tools arbitrarily without the assistance of in-hand rotation. See
Appendix~\ref{app:fixed_grasp} for implementation details.

To systematically evaluate the baselines, we set up two variations of
\texttt{sweep forward} with the \texttt{red brush} in the real-world (see
Fig.~\ref{fig:dexFuncComparison}): (i) \textit{Variation \#1} starts
with the brush top-down and does not require tool rotation, and (ii)
\textit{Variation \#2} starts with the brush sideways, which
necessitates a 90$^\circ$ tool rotation. We evaluate average
\textit{Task Progress} across 5 rollouts.

\textbf{Results.} \textit{Kinematic Retargeting} fails to grasp the
brush in both variations as kinematic motion alone does not establish
stable contacts. \textit{Fixed Grasp} succeeds on \textit{Variation
\#1} as tool rotation is not required and arm motion alone is
sufficient to finish the task. Still, it has lower \textit{Task
Progress} than \methodname\; since it is an open-loop method and
cannot react to small errors. On \textit{Variation \#2}, enforcing a
fixed grasp causes the robot arm to collide with the table while
tracking the target trajectory, as the optimizer is unable to find an
arm motion trajectory that satisfies the tool rotation while avoiding
table collision. In contrast, \methodname\; performs efficient
in-hand object rotations to complete the task.

\begin{figure}[h!]
  \centering
  \includegraphics[width=0.45\textwidth]{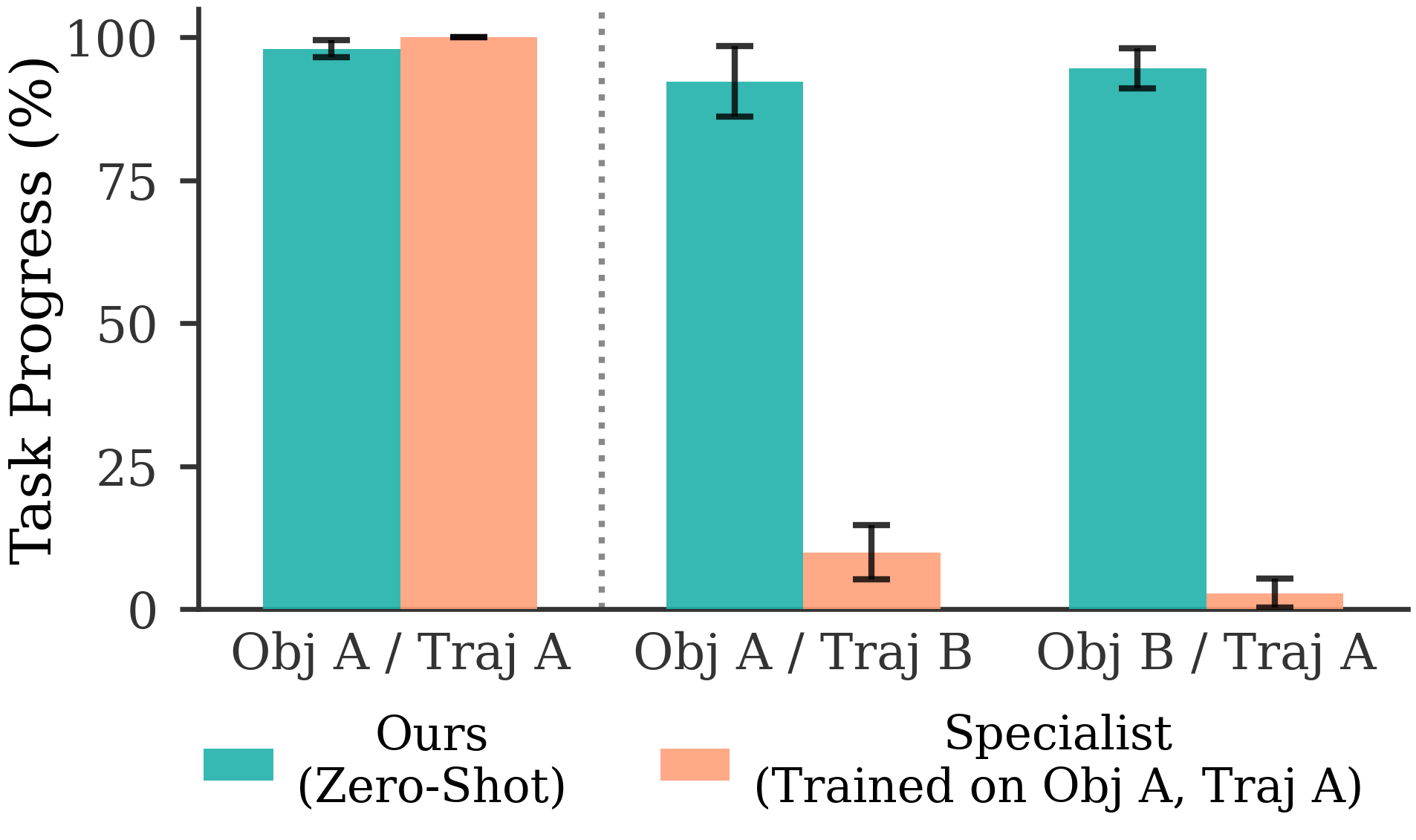}
  \captionsetup{width=0.5\textwidth}
  \caption{\textbf{Comparison against Specialists.}
    We compare \methodname\; in simulation against specialist policies
    trained on a single object (Obj A) and trajectory (Traj A). We
    train one specialist policy for each of the 6 object categories in
    \textit{DexToolBench} and report the average \textit{Task Progress}
    across these categories. While the specialists succeed on their
    training setup  (Obj A / Traj A), performance degrades under
    deviation in the trajectory (Obj A / Traj B) or the object (Obj B /
    Traj A). \methodname\; has high zero-shot performance across all
  variants, despite not being trained on these objects or trajectories.}
  \label{fig:specialistComparison}
\end{figure}

\subsection{Comparisons to Specialist Baseline}

\textbf{Setup}.
In simulation, we compare \methodname\ against task-specific
specialist RL baselines trained on a single tool and task.
Following~\cite{lum2025crossinghumanrobotembodimentgap}, we train 6
specialist policies, one for each of the 6 tool-use categories in
\textit{DexToolBench}. Each specialist policy trains on a single
object instance (Obj A) and task trajectory (Traj A). We evaluate
both methods in simulation on (i) the training setup for the
specialist: (Obj A / Traj A), (ii) the training object with a novel
trajectory (Obj A / Traj B), and (iii) a novel object instance with
the same trajectory (Obj B / Traj A). For each variation, we perform
10 rollouts for each of the 6 specialist policies, reporting the
average \textit{Task Progress} across all trials.

\textbf{Results.} Fig.~\ref{fig:specialistComparison} summarizes
these results. On the setup used for training the specialist (Obj A /
Traj A), \methodname\ matches the specialist without training on that
object or trajectory. However, the specialist’s performance drops
significantly under any variation from the training setup. When the
object stays the same, with only a change in trajectory (Obj A / Traj
B), the specialist can only track the first few goals that lift the
object. The performance is lowest when the object instance is changed
(e.g., from red brush to blue brush) even if the trajectory remains
the same (Obj B / Traj A). This indicates that the specialist
overfits to the training conditions. In contrast, \methodname\ shows
strong zero-shot \textit{Task Progress} across both object and
trajectory variations.

\subsection{Training Objective Predicts Generalization Performance}

\begin{figure}[h!]
  \includegraphics[width=0.45
  \textwidth]{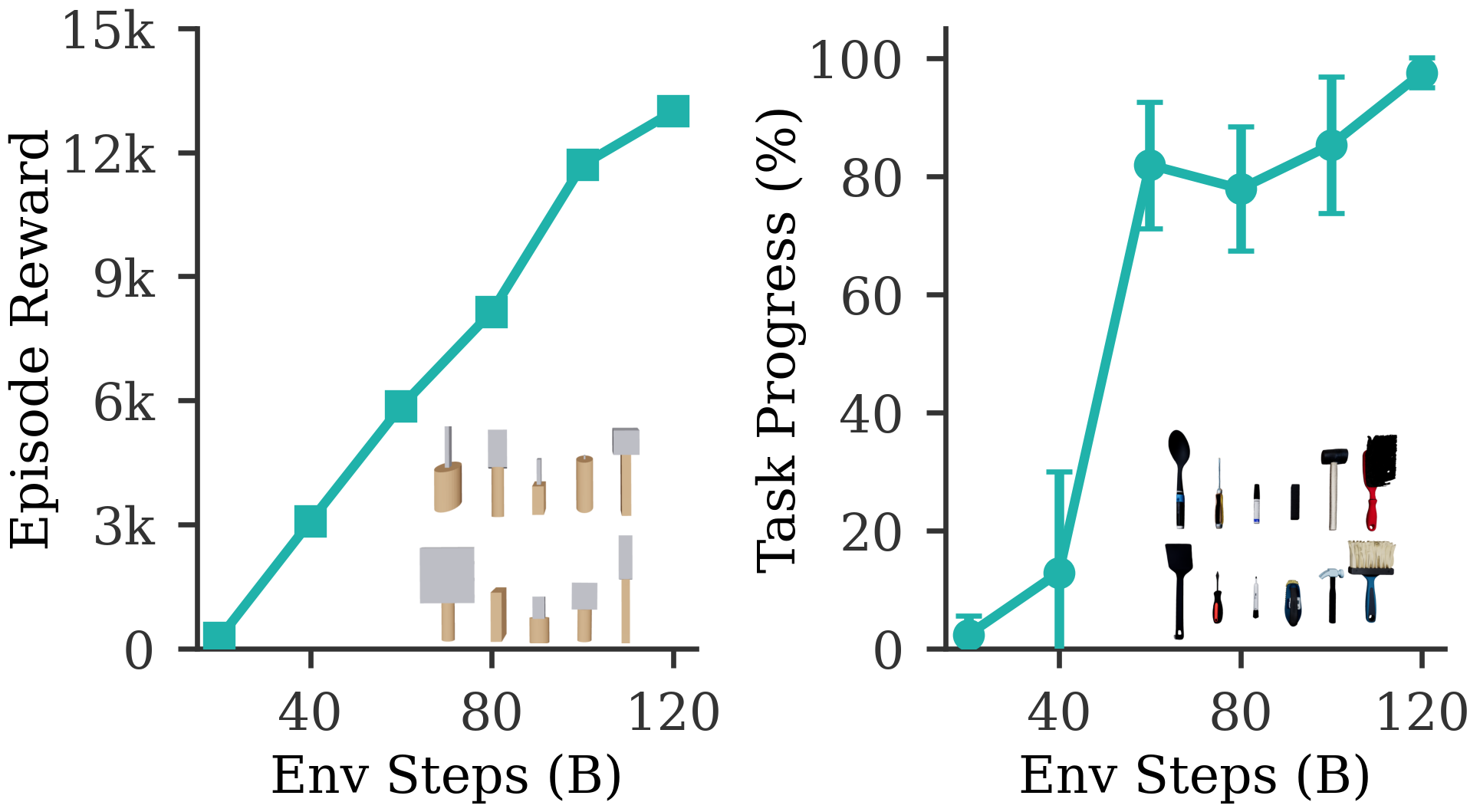}
  \captionsetup{width=0.5\textwidth}
  \caption{\textbf{Training Objective Drives Generalization.} In
    simulation, we evaluate (Left) the episode reward on
    procedurally-generated objects and (Right) the zero-shot
    \textit{Task Progress} on unseen \textit{DexToolBench} tools on
    different policy checkpoints throughout training. The strong
    correlation between the two curves validates our core hypothesis:
    improving random goal-pose reaching performance on diverse object
    primitives drives corresponding gains in generalization to unseen
  tool-use tasks.}
  \label{fig:trainTestCorrelation}
\end{figure}

\textbf{Setup.} We study how improving on our training objective:
reaching random goal poses across procedurally generated primitive
objects, relates to performance on our test objective: executing
human-demonstrated tool-use trajectories in \textit{DexToolBench}.
Fig.~\ref{fig:trainTestCorrelation} reports (i) training rewards on
the goal-pose reaching objective on primitive objects and (ii)
average \textit{Task Progress} on \textit{DexToolBench} trajectory
following tasks in simulation, plotted against the number of environment steps.

\textbf{Results.} As training proceeds, the policy’s reward on the
goal-pose reaching objective increases steadily. In parallel, the
policy’s average \textit{Task Progress} on \textit{DexToolBench}
trajectories also improves consistently. The correlation between
these curves indicates that optimizing our training objective is a
reliable predictor of downstream generalization to tool-use
behaviors, despite the mismatch between random goal-pose reaching on
primitive objects during training and trajectory following on real
tool objects at test time. This strong correlation suggests that our
randomized primitive training effectively covers the space of skills
required for real-world tool use.

\subsection{Ablations on RL Training}
\textbf{Setup.} We investigate which RL design decisions are critical
for maximizing training performance. We focus on two key components
of our RL pipeline: the choice of optimization algorithm
(SAPG~\cite{sapg2024} vs. PPO~\cite{PPO}) and the use of privileged
information (Asymmetric Critic~\cite{asymmetric_actor_critic} vs.
standard critic). We train these variants with identical
hyperparameters and environment settings, averaging results across 5
random seeds.

\begin{figure}[h!]
  \centering
  \includegraphics[width=0.45
  \textwidth]{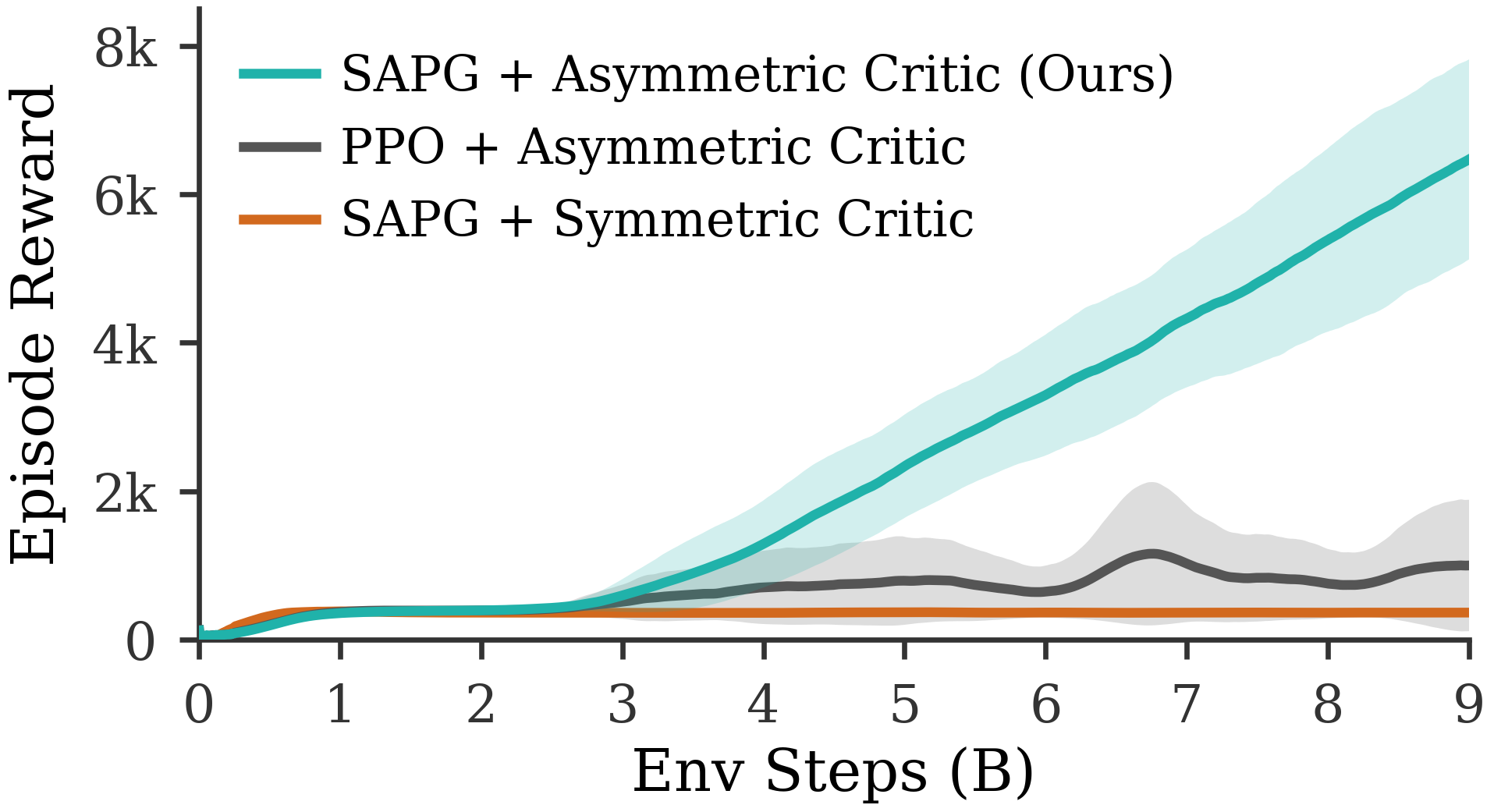}
  \captionsetup{width=0.5\textwidth}
  \caption{\textbf{Ablation of RL Training Components.} We compare
    the training reward across environment steps of \methodname\;
    against ablations averaged across 5 seeds. Replacing
    SAPG~\cite{sapg2024} with PPO~\cite{PPO} or not using Asymmetric
    Critic~\cite{asymmetric_actor_critic} results in a significant
  performance drop, highlighting their importance.}
  \label{fig:ablations}
\end{figure}

\textbf{Results.} Fig.~\ref{fig:ablations} compares the training
reward curves of our full method against these ablations.
\textit{Importance of SAPG.} We observe that replacing
SAPG~\cite{sapg2024} with PPO~\cite{PPO} leads to a significant drop
in performance. While PPO suffers from exploration saturation at
scale, SAPG mitigates this by training separate policies across
environment chunks to increase data diversity, and then fusing
gradients via importance sampling. Our results confirm that this is
essential for learning complex dexterous tool-use. \textit{Importance
of Asymmetric Critic.} We observe that removing the Asymmetric
Critic~\cite{asymmetric_actor_critic} and forcing the critic to rely
solely on the same partial observations as the actor severely hinders
learning. By accessing privileged simulation states, the critic
learns a more accurate value function to guide policy learning. This
guidance is essential for overcoming the task's partial observability.

\section{Discussion and Limitations}
We introduced \methodname, an RL framework for zero-shot dexterous
tool manipulation that generalizes to unseen tool-use tasks. Our key
insight is an object-centric reduction: tool-use can be specified as
manipulating a tool through a sequence of goal poses. This shifts the
problem from per-task reward design to a universal objective of
training goal-pose reaching policies. Training with this objective
over procedurally generated primitives induces dexterous skills, such
as stable grasping and in-hand reorientation, critical for tool-use.

\textbf{Limitations.} While our approach can track tool-use goal
sequences, it does not guarantee functional task completion,
especially for high-force interactions. Conditioning on object pose
goals alone is environment-blind, which can lead to collisions in
cluttered scenes. We also currently assume tools are rigid. Pose
alone can be insufficient to describe the state of non-rigid tools
(e.g., scissors). Finally, our high-level goal sequence is fixed and
is not replanned dynamically.

\section*{Acknowledgements}

This work is supported by Stanford Human-Centered Artificial
Intelligence (HAI), ONR Young Investigator Award, the National
Science Foundation (NSF) under Grant Numbers 2153854, 2327974,
2312956, 2327973, and 2342246, and the Natural Sciences and
Engineering Research Council of Canada (NSERC) under Award Number
526541680. We thank Sharpa for their research collaboration and for the technical support provided by their team, specifically Kaifeng Zhang, Wenjie Mei, Yi Zhou, Yunfang Yang, Jie Yin, and Jason Lee.

\bibliographystyle{plainnat}
\bibliography{references}

\clearpage

\appendix

\subsection{Reward Function Details}
\label{app:reward}

\begin{figure}[h!]
  \centering
  \includegraphics[width=0.45\textwidth]{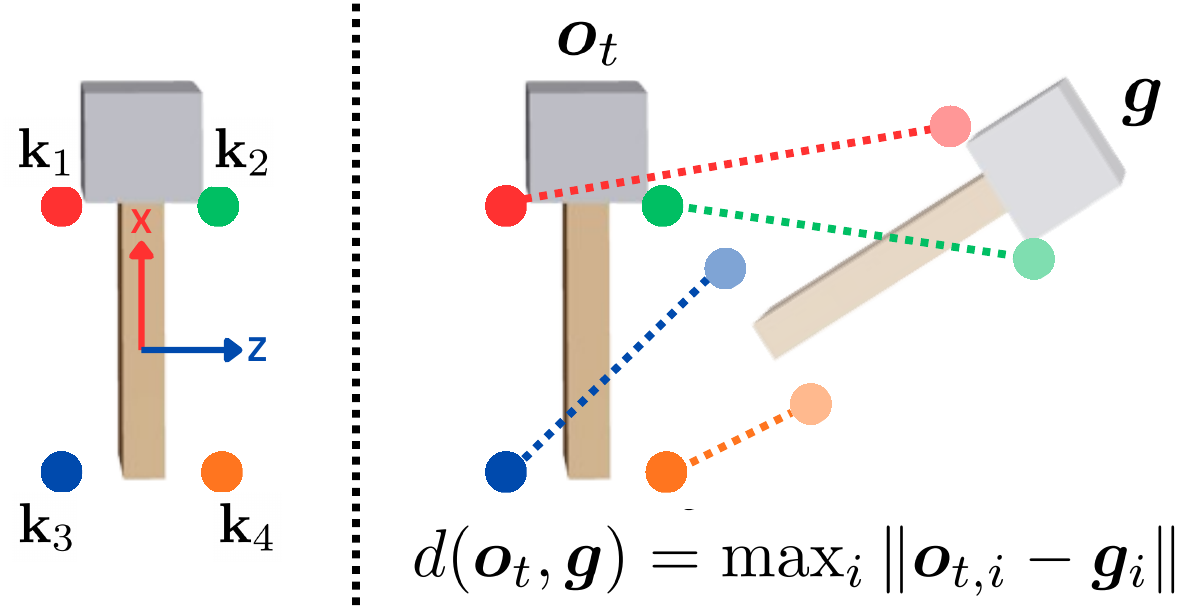}
  \captionsetup{width=0.5\textwidth}
  \caption{\textbf{Object Keypoints.} (Left) Visualization of 4 pose
    keypoints in the local object frame. (Right) Visualization of the
  keypoint distances used to compute the distance $d(\bm{o}_t, \bm{g})$.}
  \label{fig:keypoints}
\end{figure}

\begin{figure*}[t!]
  \centering
  \includegraphics[width=\textwidth]{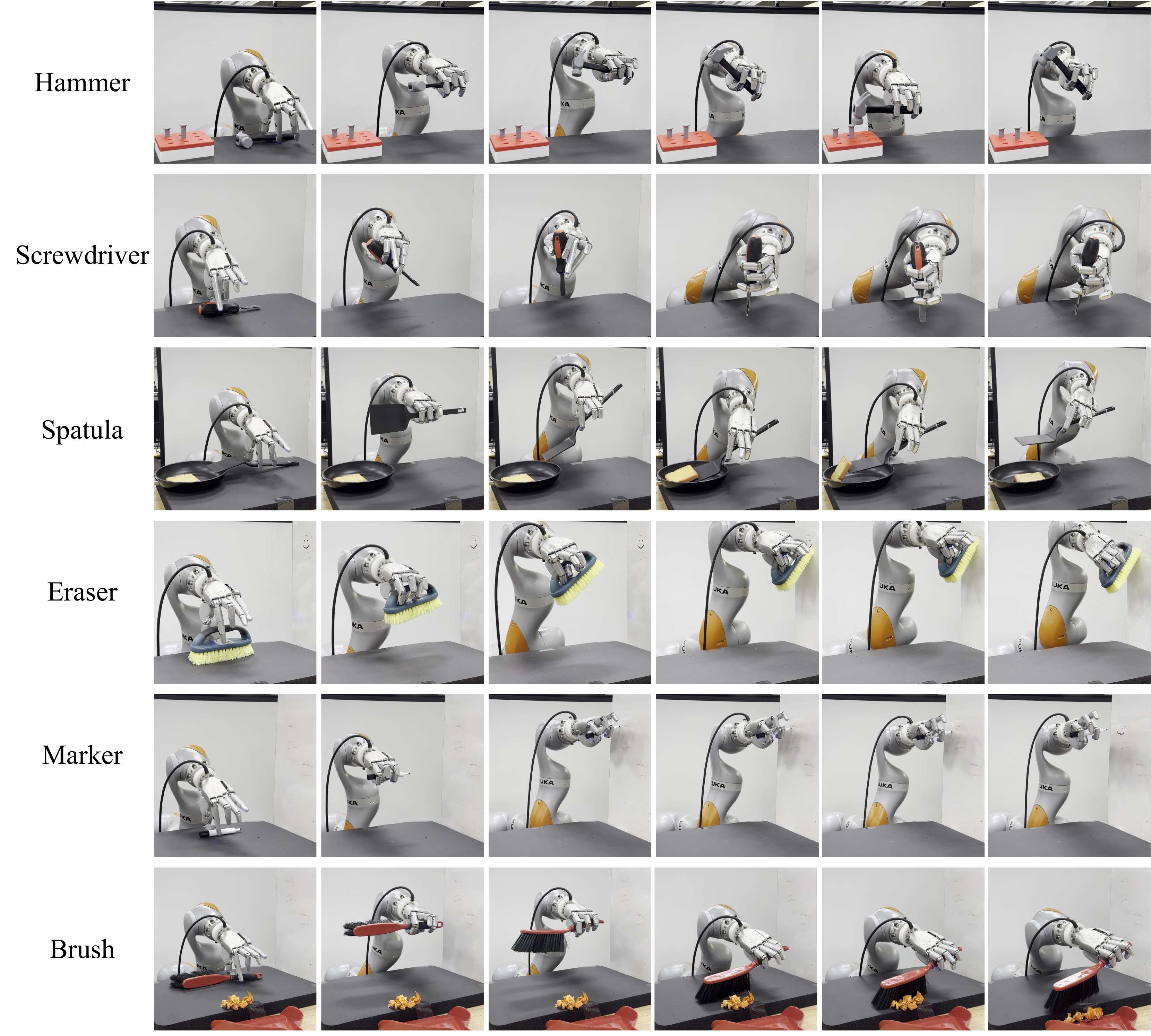}
  \captionsetup{width=\textwidth}
  \caption{\textbf{Representative Examples of \textit{DexToolBench}
    Tasks.} Visual breakdown of representative tasks in
    \textit{DexToolBench} across the 6 tool categories, highlighting
  the diversity of objects and manipulation tasks.}
  \label{fig:real_world_task_frames}
\end{figure*}

\begin{figure*}[t!]
  \centering
  \includegraphics[width=\textwidth]{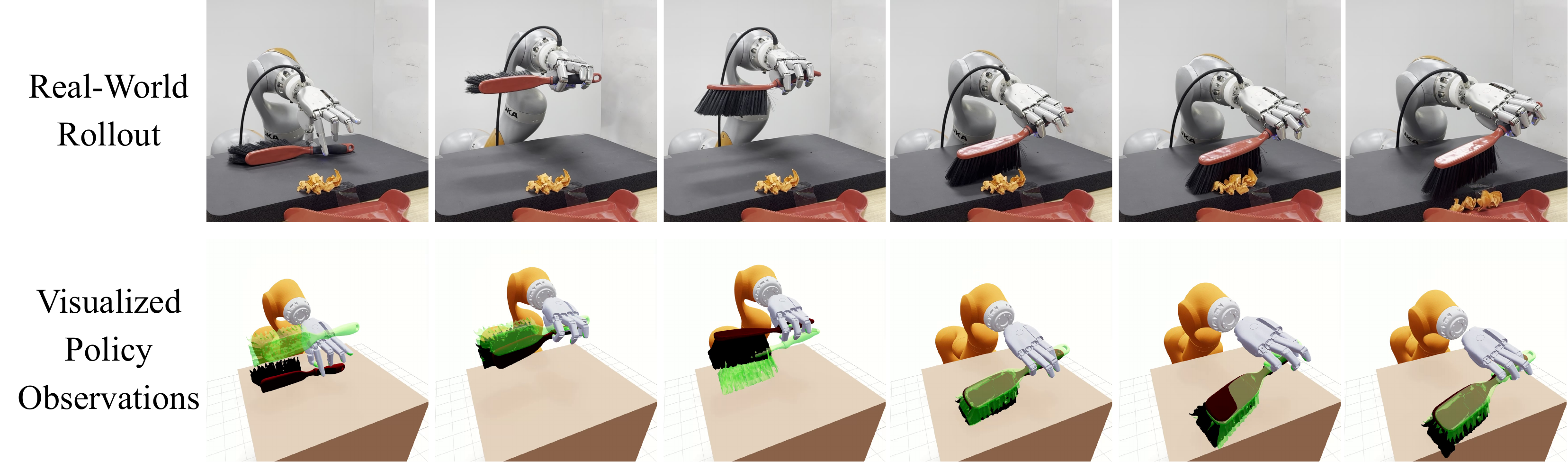}
  \captionsetup{width=\textwidth}
  \caption{\textbf{Visualization of Policy Observations during
    Real-World Deployment.}
    (Top) Image frames from a real-world rollout of the brush manipulation task.
    (Bottom) The corresponding visualization of the policy observations
    at each timestep, consisting of the robot state, estimated object
    pose, and goal pose (green). When the distance between the current
    pose and goal pose is sufficiently small, the goal pose is updated
    to the next pose in the goal sequence. Note that this visualization
  is a rendering of the policy's inputs, not a physics-based simulation.}
  \label{fig:visualized_policy_observations}
\end{figure*}

\section{Reward Function Details}
\label{app:reward}
The reward function $r$ has three primary components:
\begin{equation}
  r = r_{\text{smooth}} + r_{\text{grasp}} +
  \mathbb{I}_{\text{grasped}}r_{\text{goal}}
\end{equation}
where $\mathbb{I}_{\text{grasped}}$ is an indicator function that
triggers once the object has been successfully grasped. We describe
each reward term in further detail below:

\textbf{Smoothness Reward ($r_{\text{smooth}}$).} To promote
physically plausible control and reduce hardware wear, we penalize
the $L_1$ norm of joint velocities:
\begin{equation}
  r_{\text{smooth}} = -\lambda_{\text{arm}}
  \|\dot{\mathbf{q}}_{\text{arm}}\|_1 - \lambda_{\text{hand}}
  \|\dot{\mathbf{q}}_{\text{hand}}\|_1
\end{equation}
where $\dot{\mathbf{q}}^{\text{arm}}$ and
$\dot{\mathbf{q}}^{\text{hand}}$ correspond to the current velocities
of the 7-DOF Kuka arm and the 22-DoF Sharpa hand, respectively.

\textbf{Grasp Reward ($r_{\text{grasp}}$).} The term
$r_{\text{grasp}}$ facilitates the transition from a neutral pose to
a stable grasp:
\begin{equation}
  r_{\text{grasp}} = r_{\text{approach}} + (1 -
  \mathbb{I}_{\text{grasped}})r_{\text{lift}}
\end{equation}

$r_{\text{approach}}$ encourages the agent to reduce the distance
between the robot's hand and the object. It rewards the agent for
approaching the object rather than simply maintaining a fixed close
distance. We define:
\begin{equation}
  r_{\text{approach}} = \lambda_{\text{approach}}
  \max(\bar{d}^{*}_{\text{ft}} - \bar{d}_\text{ft}, 0)
\end{equation}
where $\bar{d}_\text{ft}$ is the current mean distance between the
fingertips and the object, and $\bar{d}^{*}_{\text{ft}}$ is a
stateful variable that stores the minimum mean distance achieved so
far in the episode. $r_{\text{lift}}$ encourages the agent to grasp
and lift the object. We define:
\begin{equation}
  r_{\text{lift}} = \lambda_{\text{lift}} \max(z - z_{\text{init}},
  0) + \mathbb{I}[z \ge z_{\text{lifted}}] B_{\text{lifted}}
\end{equation}
where $z$ is the vertical position of the object, $z_{\text{init}}$
is the initial $z$ position of the object, $z_{\text{lifted}}$ is the
lifted threshold, and $B_{\text{lifted}}$ is a bonus that is awarded
at most once per episode when the object has been lifted.
$\mathbb{I}_{\text{grasped}}$ turns true once $z \ge
z_{\text{lifted}}$. This ensures that this lifting reward only
applies at the start of the episode before the object has been
lifted, and then is 0 for the remainder of the episode. At this
point, $r_{\text{goal}}$ takes over.

\textbf{Goal-Pose Reward ($r_{\text{goal}}$).} Once
$\mathbb{I}_{\text{grasped}} = 1$, the goal-reaching term
$r_{\text{goal}}$ becomes the dominant signal to train our RL policy.
As Section~\ref{sec:training} describes, we define:
\begin{equation}
  r_{\rm goal}
  = \textrm{max}(d^* - d(\bm{o}_t,\bm{g}), 0) + B_{\rm
  succ}\,\mathbb{I}\!\left[d(\bm{o}_t,\bm{g}) < \epsilon\right],
\end{equation}

Our goal-reaching reward has (i) a dense progress term based on the
goal distance $d(\bm{o}_t,\bm{g})$ and (ii) a sparse success bonus
$B_{\rm succ}$. We maintain a stateful variable $d^*$ storing the
smallest distance achieved for the current goal (reset when a new
goal is sampled), and only reward improvements in distance to avoid
incentivizing static proximity. When the goal is reached, i.e.,
$d(\bm{o}_t,\bm{g})<\epsilon$, the agent receives $B_{\rm succ}$ and
we resample a new goal.

\textit{Keypoint Distance Formulation:}
Following~\cite{allshire2022transferring,petrenko2023dexpbt}, we
measure distance using $D{=}4$ object-frame keypoints $
d(\bm{o}_t,\bm{g}) = \max_i \left\| \bm{o}_{t,i} - \bm{g}_i \right\|,$
where $\bm{o}_{t,i}$ and $\bm{g}_i$ are the world-frame positions of
the $i$-th keypoint under the current and goal poses, respectively.
We define the following scales $\mathbf{s}^{\text{rew}} =
[s_x^{\text{rew}}, s_y^{\text{rew}}, s_z^{\text{rew}}] = [0.14, 0.03,
0.03]$ (in metres), and we define the keypoints
$\{\mathbf{k}_i\}_{i=1}^4$ at the following offsets:
\begin{equation}
  \mathbf{k} \in \left\{
    \begin{bmatrix} s_x^{\text{rew}}/2 \\ s_y^{\text{rew}}/2
      \\ s_z^{\text{rew}}/2
    \end{bmatrix},
    \begin{bmatrix} s_x^{\text{rew}}/2 \\ s_y^{\text{rew}}/2
      \\ -s_z^{\text{rew}}/2
    \end{bmatrix},
    \begin{bmatrix} -s_x^{\text{rew}}/2 \\ -s_y^{\text{rew}}/2
      \\ s_z^{\text{rew}}/2
    \end{bmatrix},
    \begin{bmatrix} -s_x^{\text{rew}}/2 \\ -s_y^{\text{rew}}/2
      \\ -s_z^{\text{rew}}/2
    \end{bmatrix}
  \right\}
\end{equation}

Fig.~\ref{fig:keypoints} visualizes our keypoint-based pose
representation. Defining pose distance in this space yields a single,
interpretable metric that jointly captures translation and rotation
error. Since many tools are elongated, we set $s_x^{\text{rew}} >
s_y^{\text{rew}}, s_z^{\text{rew}}$ so the reward is more sensitive
to pitch and yaw errors (rotations about the $y$ and $z$ axes) than
to roll about the long $x$-axis. This biases the policy toward
aligning the tool’s principal axis with the target pose, which is
typically the dominant requirement for tool use. Using fixed relative
scales across objects also ensures a consistent trade-off between
translational and rotational progress for all tasks.

\subsection{Procedural Asset Generation Details}
\label{app:asset_gen}
To promote robustness to a wide range of inertial properties, we
procedurally generate tool-like objects using a simple
\emph{handle--head} abstraction. This minimal design still spans a
broad family of real-world handheld tools (e.g., brushes, markers,
spatulas, screwdrivers, and hammers). The goal is to expose the
policy to large variations in geometry and physics without relying on
complex meshes that slow down simulation and training. Each tool is
composed of cuboids and cylinder primitives, which ensures stable
physics and fast simulation.

\textbf{Tool parameterization.}
Each asset consists of (i) a handle and (ii) a head rigidly attached
to one end of the handle. We randomize both geometry and density. For
each part, we sample its length $L$ and cross-sectional dimensions.
For cuboids we use width $W$ and height $H$; for cylindrical/capsule
variants, $W$ denotes the diameter and we set $H=W$.

\textbf{Geometric randomizations.}
Each tool is a rigid union of two parts: a \emph{handle} and a
\emph{head}. For each part, we randomly choose one of two primitive
shapes: a \emph{cuboid} (box) or a \emph{capsule} (cylinder with
rounded ends). We then sample the part dimensions uniformly from
ranges chosen to span common handheld tools (from small
brushes/markers to larger hammers). The head is attached to one end
of the handle, and the head's long axis is rotated by $90^\circ$
relative to the handle's long axis. For a \emph{cuboid}, we sample a
length, width, and height. For a \emph{capsule}, we sample a length
and a diameter:

\emph{Handle dimensions:} length is sampled in $[5,30]$\,cm;
width/height (or diameter) is sampled in $[1,4]$\,cm.

\emph{Head dimensions:} length is sampled in $[1,15]$\,cm;
width/height (or diameter) is sampled in $[0.5,12]$\,cm.

\textbf{Physics randomizations.}
To systematically induce physics randomizations, we assign different
densities to the handle and head. Handles are sampled with a lower density,
$\rho_{\text{low}} \sim \mathcal{U}[300,600]\ \text{kg/m}^3$,
approximating wood or plastics. Heads are sampled with a higher density range
$\rho_{\text{high}} \sim \mathcal{U}[300,2000]\ \text{kg/m}^3$,
approximating metals or dense rubber. Together with geometric
variation, this density variation results in diverse center-of-mass locations
(often shifted toward the head) and a broad range of rotational
inertias, encouraging the policy to learn to adapt under varying
physics in the real-world.

\subsection{Simulation Training Details}

\subsubsection{Simulation Details} We use
IsaacGym~\cite{makoviychuk2021isaacgym} for massively-parallel
GPU-accelerated simulation. We build from the DexPBT Kuka Allegro
Reorientation environment~\cite{petrenko2023dexpbt}, and add
substantial changes to the environment to improve dexterous tool-use,
enable sim-to-real transfer, and improve generalization to unseen
objects and trajectories.

Notably, many other GPU-accelerated simulators, such as
Genesis~\cite{Genesis} and MJWarp~\cite{mujoco_playground_2025},
could not be used because they do not currently support simulation of
parallel environments with different object geometries per scene.

\subsubsection{Policy Action}
The RL policy outputs an action $\bm{a}_t \in \mathbb{R}^{J}$, where
$J=29$ represents the total number of actuated joints. We first clip
the actions to the range $[-1, 1]$. We partition this action into arm
components $\bm{a}^{\text{arm}}_t \in [-1, 1]^{7}$ and hand
components $\bm{a}^{\text{hand}}_t \in [-1, 1]^{22}$. These are
processed into joint position targets $\bm{q}^{\text{target}}_t$ as follows:

\textbf{Arm Control (Delta).} For the 7-DoF arm, we interpret the
action as a relative displacement (delta) from the previous target.
The intermediate target is computed as:
\begin{equation}
  \hat{\bm{q}}^{\text{arm}}_t = \bm{q}^{\text{target, arm}}_{t-1} +
  k^{\text{arm}} \bm{a}^{\text{arm}}_t
\end{equation}
where $k^{\text{arm}} = 0.025$ is a scaling factor derived from the
control frequency and speed limits. We clip
$\hat{\bm{q}}^{\text{arm}}_t$ to the joint limits, and then apply an
exponential moving average (EMA) filter with smoothing factor
$\alpha_{\text{arm}}$ to obtain the final $\bm{q}^{\text{target, arm}}_t$.

\textbf{Hand Control (Absolute).} For the 22-DoF hand, we interpret
the action as an absolute target within the joint limits. We map the
action range $[-1, 1]$ to the physical limits
$[\bm{q}^{\text{hand}}_{\text{lower}},
\bm{q}^{\text{hand}}_{\text{upper}}]$ via an affine transformation:
\begin{equation}
  \hat{\bm{q}}^{\text{hand}}_t = \frac{\bm{a}^{\text{hand}}_t + 1}{2}
  \odot (\bm{q}^{\text{hand}}_{\text{upper}} -
  \bm{q}^{\text{hand}}_{\text{lower}}) + \bm{q}^{\text{hand}}_{\text{lower}}
\end{equation}
We then apply an EMA filter with smoothing factor
$\alpha_{\text{hand}}$ to $\hat{\bm{q}}^{\text{hand}}_t$, and finally
clip the result to the joint limits to ensure safety, yielding
$\bm{q}^{\text{target, hand}}_t$.

In our experiments, we use $\alpha_{\text{arm}} =
\alpha_{\text{hand}} = 0.1$, where $\alpha=1.0$ means no smoothing.

\subsubsection{Policy Observation}
The policy observation space $\mathcal{O}$ comprises robot
proprioception $\bm{s}_t$, object state $\bm{o}_t$, goal information
$\bm{g}$, and the object descriptor $\bm{\phi}$. We typically provide
relative representations (e.g., keypoints expressed relative to the
palm) over absolute world positions to facilitate generalization
across tool geometries and improve invariance to absolute world
coordinates. The specific observations are:

\textbf{Proprioception ($\bm{s}_t$).}
\begin{itemize}
  \item Joint positions $\bm{q}_t \in \mathbb{R}^{29}$ and velocities
    $\dot{\bm{q}}_t \in \mathbb{R}^{29}$.
  \item Previous joint position targets $\bm{q}^{\text{target}}_{t-1}
    \in \mathbb{R}^{29}$.
  \item Palm pose, consisting of world position $\bm{x}_{\text{palm}}
    \in \mathbb{R}^3$ and orientation quaternion
    $\bm{r}_{\text{palm}} \in \mathbb{R}^4$.
  \item Fingertip positions relative to the palm center:
    $\{\bm{x}_{\text{tip}, j} - \bm{x}_{\text{palm}}\}_{j=1}^5 \in
    \mathbb{R}^{15}$.
\end{itemize}

\textbf{Object Pose, Goal Pose, and Descriptor ($\bm{o}_t, \bm{g}, \bm{\phi}$).}
Instead of directly inputting raw pose matrices, we represent the
object and goal states using $K=4$ keypoints. These observation
keypoints are computed using the object's grasp bounding box $\bm{s}
\in \mathbb{R}^3$ (length, width, height), effectively conditioning
the policy on the specific tool geometry.
\begin{itemize}
  \item Object orientation quaternion $\bm{r}_{\text{obj}} \in \mathbb{R}^4$.
  \item Object keypoints relative to the palm: $\{\bm{o}_{t,i} -
    \bm{x}_{\text{palm}}\}_{i=1}^4 \in \mathbb{R}^{12}$.
  \item Keypoint errors to goal: $\{\bm{o}_{t,i} -
    \bm{g}_{i}\}_{i=1}^4 \in \mathbb{R}^{12}$. Here, $\bm{o}_{t,i}$
    and $\bm{g}_i$ correspond to the $i$-th keypoint of the current
    and goal poses, respectively.
  \item Object scales $\bm{s} \in \mathbb{R}^3$, representing the
    dimensions of the tool's graspable region (handle).
\end{itemize}

We note that while the reward function (Appendix~\ref{app:reward})
uses \textit{fixed} keypoint offsets to maintain a consistent success
metric across objects, the policy observations use the
\textit{instance-specific} offsets defined by the object scales
$\bm{s} = [s_x, s_y, s_z]$:
\begin{equation}
  \mathbf{k}_{\text{obs}} \in \left\{
    \begin{bmatrix} s_x/2 \\ s_y/2 \\ s_z/2
    \end{bmatrix},
    \begin{bmatrix} s_x/2 \\ s_y/2 \\ -s_z/2
    \end{bmatrix},
    \begin{bmatrix} -s_x/2 \\ -s_y/2 \\ s_z/2
    \end{bmatrix},
    \begin{bmatrix} -s_x/2 \\ -s_y/2 \\ -s_z/2
    \end{bmatrix}
  \right\}
\end{equation}

\subsubsection{Asymmetric Critic State}
We utilize an asymmetric actor-critic setup where the critic is
trained on a privileged state space $\mathcal{S}$, while the policy
acts on the restricted observation space $\mathcal{O}$. The critic
state $\mathcal{S}$ contains the exact, noise-free, and instantaneous
ground-truth state of the system, whereas the policy observation
$\mathcal{O}$ is subjected to noise and delays to bridge the
sim-to-real gap. In addition to the standard observations, the critic
state includes:
\begin{itemize}
  \item Ground-truth Velocities: Exact linear and angular velocities
    for the palm ($\bm{v}_{\text{palm}}, \bm{\omega}_{\text{palm}}$)
    and the object ($\bm{v}_{\text{obj}}, \bm{\omega}_{\text{obj}}$).
  \item Reward Signals: The instantaneous reward $r_t$ and the number
    of successes achieved this episode, which help to improve value
    function estimation.
  \item Stateful Progress Features: Auxiliary features including the
    minimum fingertip-to-object distance achieved since the last
    reset, the minimum keypoint distance achieved since last reset,
    the number of environment steps that have occurred since last
    reset, and a binary signal $\mathbb{I}_{\text{grasped}}$
    indicating if the object has been lifted.
  \item Noise-Free and Undelayed Object Pose: Exact object pose
    without noise or delays.
\end{itemize}

\subsubsection{Reset and Initialization} Next, we describe the
termination conditions and initialization of the environment.

\textbf{Episode Termination Conditions.} An environment is reset if
any of the following conditions are met:
\begin{itemize}
  \item Object Fallen Off Table: The object height drops below the
    table surface.
  \item Object Drop (Hysteresis): To penalize dropping the object
    after grasping, we trigger a reset if the object returns to the
    table surface ($z < z_{\text{init}}$) \textit{after} the grasped
    flag $\mathbb{I}_{\text{grasped}}$ has been activated.
  \item Hand Wander: The distance between the hand and the object
    exceeds $1.5$\,m.
  \item Table Force Limits: The net force measured by the table
    sensor exceeds $100$\,N, preventing aggressive collisions.
  \item Timeout: The episode duration exceeds the maximum step limit.
  \item Max Consecutive Successes: The agent achieves the maximum
    number of consecutive successes for an episode.
\end{itemize}

\textbf{Scene Initialization.}
At the start of each episode, the robot and object are reset to
randomized states to encourage robustness:
\begin{itemize}
  \item Robot: Joint positions are initialized to a default position
    with additive uniform noise.
  \item Object: The object is spawned above the table surface with a
    randomized planar position $(x, y)$ and a random rotation
    quaternion $\bm{r} \in SO(3)$. We also apply small random
    perturbations to the table height to improve robustness to
    geometric calibration errors.
\end{itemize}

\textbf{Goal Sampling Strategy.}
\begin{enumerate}
  \item Initial Goal ($\bm{g}_0$): The first goal is sampled
    uniformly from a 3D workspace volume $V$ centered above the
    table. The sampling ranges relative to the table center are $x
    \in [-0.35, 0.35]$, $y \in [-0.1, 0.2]$, and $z \in [0.15,
    0.52]$, where $z$ is up, $-y$ is forward (wrt the robot), and $x$
    is left (wrt the robot).
  \item Subsequent Goals ($\bm{g}_{k+1}$): Upon successfully reaching
    goal $\bm{g}_k$, we sample the next goal relative to the
    \textit{previous goal}. We apply a random perturbation of up to
    $0.1$\,m in position and $90^\circ$ in rotation.
\end{enumerate}

\begin{table*}[h]
  \centering
  \caption{Simulation environment and SAPG training hyperparameters.}
  \label{tab:hyperparameters}
  \resizebox{\textwidth}{!}{%
    \begin{tabular}{@{}lc@{\hskip 0.5in}lc@{}}
      \toprule
      \textbf{Parameter} & \textbf{Value} & \textbf{Parameter} &
      \textbf{Value} \\ \midrule
      \multicolumn{2}{@{}l}{\textit{\textbf{Environment \& Control}}}
      & \multicolumn{2}{@{}l}{\textit{\textbf{SAPG Hyperparameters}}} \\
      Simulation / Control Frequency & 120 / 60 Hz & Actor Network &
      LSTM[1024] + MLP[1024,1024,512,512] \\
      Num. Environments & 24576 & Critic Network & MLP[1024,1024,512,512] \\
      Episode Length & 600 steps & Learning Rate & $1 \times 10^{-4}$ \\
      Obj. Pos. Range ($x,y$) & $\pm\, 10$ cm & Minibatch Size & 98,304 \\
      Table Height Range ($z$) & $\pm\, 1$ cm & SAPG Block Size & 4096 \\
      Robot Joint Pos. Range & $\pm\, 0.1$ rad & Entropy Bonus Scale & 0.005 \\
      Success Tolerance ($\epsilon$) & 1 cm & Discount Factor
      ($\gamma$) & 0.99 \\
      Initial Height ($z_{\text{init}}$) & 0.63 m & GAE Parameter
      ($\tau$) & 0.95 \\
      Lift Threshold ($z_{\text{lifted}}$) & 0.73 m & Clip Range & 0.1 \\
      & & & \\
      \multicolumn{2}{@{}l}{\textit{\textbf{Domain Randomization}}} &
      \multicolumn{2}{@{}l}{\textit{\textbf{Reward Coefficients}}} \\
      Obj. Pose Noise (Trans.) & 1 cm & Arm Action Penalty
      ($\lambda_{\text{arm}}$) & 0.03 \\
      Obj. Pose Noise (Rot.) & $5.0^\circ$ & Hand Action Penalty
      ($\lambda_{\text{hand}}$) & 0.003 \\
      Obj. Pose Delay Max & 10 steps & Approach Scale
      ($\lambda_{\text{approach}}$) & 50.0 \\
      Action/Obs Delay Max & 3 steps & Lifting Scale
      ($\lambda_{\text{lift}}$) & 20.0 \\
      Joint Vel. Obs Noise ($\sigma$) & 0.1 rad/s & Lifting Bonus
      ($B_{\text{lifted}}$) & 300.0 \\
      Perturb. Force Scale & 5.0 N & Goal Pose Scale
      ($\lambda_{\text{goal}}$) & 200.0 \\
      Perturb. Torque Scale & 0.5 Nm & Success Bonus
      ($B_{\text{succ}})$ & 1000.0 \\ \bottomrule
    \end{tabular}%
  }
\end{table*}

\subsubsection{Hyperparameters} Table~\ref{tab:hyperparameters} shows
the important hyperparameters used for the simulation environment and
SAPG training.

\subsection{Sim-to-Real Details}
\label{app:sim-to-real}

\begin{itemize}
  \item Observation and action delay: We model system latency by
    passing observations, actions, and object states through FIFO
    queues and randomly sampling delayed values. To reflect the
    uncertainty and inference latency of vision-based object pose
    tracking, object pose estimates are assigned a significantly
    larger delay than other signals.
  \item Accurate robot modeling: We explicitly match the simulation's
    joint gains, armature, and damping to those of the physical robot.
  \item Sensor noise: We inject zero-mean Gaussian noise into joint
    velocities and object pose observations, as we found these
    signals to exhibit the highest measurement noise in the real
    world. The noise magnitudes are calibrated to match empirical
    levels, robustifying the policy against realistic sensor noise.
  \item Smooth actions: We apply exponential moving average (EMA)
    smoothing to joint position targets to prevent high-frequency
    control artifacts.
  \item Table height randomization: We randomize the table height at
    the start of each episode to improve robustness to geometric
    variations in the physical workspace.
  \item External force and torque perturbations: We apply random
    external forces and torques to the object during training to help
    the policy generalize to unmodeled contact dynamics and
    unexpected disturbances.
  \item Finger self-collisions: We restrict the range of the finger
    abduction/adduction joints to a conservative subset of their
    physical limits. This prevents the policy from reaching
    configurations that could result in potentially damaging
    self-collisions on the real hardware.
\end{itemize}

\subsection{Human Video Processing Details}
\label{app:human_video_processing}

\begin{figure*}[h!]
  \centering
  \includegraphics[width=\textwidth]{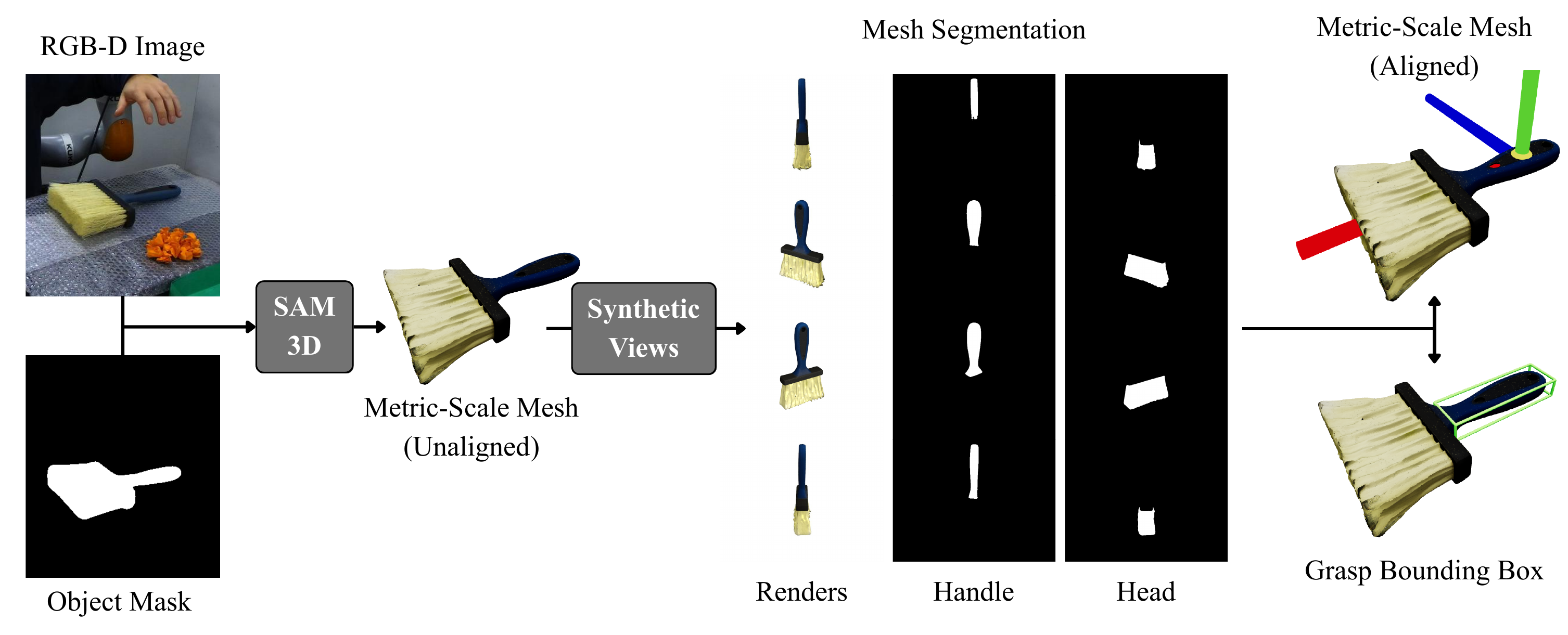}
  \captionsetup{width=\textwidth}
  \caption{\textbf{Metric-Scale Mesh and Grasp Bounding Box Acquisition.}
    We present a semi-automated pipeline to extract a metric-scale
    mesh and grasp bounding box from a single RGB-D video scan.
    (1) We first segment the target object in the initial frame and
    reconstruct a 3D mesh using SAM 3D~\cite{chen2025sam}, injecting
    the captured depth map to ensure metric accuracy.
    (2) To establish a canonical coordinate frame, we virtually
    render the mesh from multiple views and use SAM
    2~\cite{ravi2024sam2} to segment the geometry into ``handle'' and
    ``head'' regions based on user prompts.
    (3) The final grasp bounding box is derived from the handle's
    geometry: it is centered on the handle, with the positive $x$-axis
  oriented toward the head, ensuring consistent alignment with policy training.}
  \label{fig:sam_pipeline}
\end{figure*}

\subsubsection{Metric-Scale Mesh and Grasp Bounding Box Acquisition}
Fig.~\ref{fig:sam_pipeline} illustrates our pipeline, which leverages
SAM 3D~\cite{chen2025sam} to extract a metric-scale mesh and grasp
bounding box from a single RGB-D video demonstration.

\begin{enumerate}[label=\alph*)]
\item \textbf{Initial Object Segmentation}: We first obtain a
  segmentation mask of the object from the first frame of the video.
  We use the first frame because it typically minimizes occlusion, as
  this is prior to human interaction. Although segmentation could be
  conditioned on a text prompt, we found that prompting SAM
  2~\cite{ravi2024sam2} with two user-specified points on the object
  provided superior reliability.
\item \textbf{Metric-Scale Mesh Extraction}: Next, we generate a 3D
  mesh using SAM 3D~\cite{chen2025sam}. Although the default SAM 3D
  pipeline relies on monocular depth estimation from RGB input, the
  resulting mesh is not metrically accurate. Since we require
  accurate scale for both pose tracking and grasp bounding box
  estimation, we modify the pipeline to directly utilize the captured
  depth image instead of the predicted depth.
\item \textbf{Part Segmentation (Handle vs. Head)}: To define the
  object's origin, we segment the mesh into a \textit{handle}
  (graspable region) and a \textit{head} (non-graspable region). This
  decomposition allows us to center the grasp bounding box on the
  handle and align the object's x-axis along the handle's primary
  axis (x-axis pointing from handle to head). We automate this by
  rendering a video of the mesh from a virtual camera circling the
  object. The user provides point prompts on the first frame to
  define the handle and head, and we use SAM 2~\cite{ravi2024sam2} to
  propagate these masks across the rendered sequence.
\item \textbf{Grasp Bounding Box Definition}: Using the rendered
  camera extrinsics and intrinsics, we back-project the masked depth
  maps to 3D to obtain separate point clouds for the handle and head.
  We crop the full mesh based on these point clouds. Finally, we
  define the grasp bounding box using the handle's geometry: the box
  is centered at the handle's centroid, with its x-axis oriented
  along the vector pointing from the handle centroid toward the head centroid.
\end{enumerate}

\subsubsection{Goal Pose Sequence}

We use FoundationPose to extract the raw object pose trajectory from
the RGB-D human video demonstration, which is collected at the
camera's native frequency of 30Hz. Directly tracking this raw
sequence is suboptimal for two reasons: (1) frame-to-frame pose
estimation jitter can make the goal pose sequence shaky, and (2) the
demonstration often includes a pre-grasp phase where the object
remains stationary on the table while the human approaches it. To
address these issues, we apply the following preprocessing steps.

\textbf{Temporal Downsampling.} We downsample the trajectory from
30Hz to 3Hz. This acts as a low-pass filter, removing high-frequency
perception noise and ensuring the policy tracks the underlying smooth
motion profile rather than attempting to replicate spurious artifacts
in the pose estimation.

\textbf{Lift-off Truncation.} We automatically trim the start of the
trajectory to remove the static phase where the object rests on the
table. We calculate the object's vertical position $z_t$ relative to
the table surface at each timestep. The goal sequence is defined to
start at the first frame $t_{\text{start}}$ where the object's height
exceeds a threshold $z_{\text{thresh}} = 10\text{cm}$. This ensures
that the robot immediately attempts to lift and manipulate the tool,
rather than trying to maintain a static pose on the table surface.

\begin{table*}[h]
  \centering
  \caption{\textbf{Detailed Real-World Evaluation Results.} We report
    the \textit{Task Progress} (\%) for each of the 5 rollouts across
    all 24 object-task variations. The specific tools correspond to the
  instances shown in Fig.~\ref{fig:generalizationResults}.}
  \label{tab:detailed_results}
  \resizebox{0.7\textwidth}{!}{%
    \begin{tabular}{lllccccc|c}
      \toprule
      \textbf{Category} & \textbf{Instance} & \textbf{Trajectory} &
      \textbf{R1} & \textbf{R2} & \textbf{R3} & \textbf{R4} &
      \textbf{R5} & \textbf{Avg} \\
      \midrule
      \multirow{4}{*}{Hammer} & \multirow{2}{*}{Claw} & Swing Down &
      100 & 100 & 100 & 100 & 100 & 100 \\
      &  & Swing Side & 100 & 100 & 77.5 & 100 & 100 & 95.5 \\
      \cmidrule{2-9}
      & \multirow{2}{*}{Mallet} & Swing Down & 100 & 100 & 33.3 & 100
      & 88.9 & 84.4 \\
      &  & Swing Side & 65.6 & 81.3 & 100 & 78.1 & 62.5 & 77.5 \\
      \midrule
      \multirow{4}{*}{Marker} & \multirow{2}{*}{Sharpie} & Draw Smile
      & 100 & 100 & 100 & 100 & 0.0 & 80.0 \\
      &  & Write C & 100 & 100 & 100 & 32.0 & 56.0 & 77.6 \\
      \cmidrule{2-9}
      & \multirow{2}{*}{Staples} & Draw Smile & 100 & 53.1 & 100 &
      100 & 100 & 90.6 \\
      &  & Write C & 31.0 & 0.0 & 100 & 100 & 100 & 66.2 \\
      \midrule
      \multirow{4}{*}{Eraser} & \multirow{2}{*}{Handle} & Wipe Smile
      & 100 & 100 & 100 & 100 & 100 & 100 \\
      &  & Wipe C & 100 & 100 & 100 & 100 & 100 & 100 \\
      \cmidrule{2-9}
      & \multirow{2}{*}{Flat} & Wipe Smile & 100 & 100 & 100 & 100 &
      100 & 100 \\
      &  & Wipe C & 100 & 100 & 100 & 100 & 100 & 100 \\
      \midrule
      \multirow{4}{*}{Brush} & \multirow{2}{*}{Blue} & Sweep Fwd &
      47.8 & 47.8 & 40.0 & 60.0 & 60.0 & 51.1 \\
      &  & Sweep Right & 100 & 100 & 100 & 100 & 100 & 100 \\
      \cmidrule{2-9}
      & \multirow{2}{*}{Red} & Sweep Fwd & 100 & 100 & 13.5 & 100 &
      100 & 82.7 \\
      &  & Sweep Right & 43.2 & 62.2 & 100 & 42.2 & 59.5 & 61.4 \\
      \midrule
      \multirow{4}{*}{Spatula} & \multirow{2}{*}{Spoon} & Serve Plate
      & 100 & 76.5 & 100 & 48.5 & 100 & 85.0 \\
      &  & Flip Over & 77.5 & 100 & 100 & 100 & 100 & 95.5 \\
      \cmidrule{2-9}
      & \multirow{2}{*}{Flat} & Serve Plate & 100 & 75.0 & 78.1 &
      56.3 & 78.1 & 77.5 \\
      &  & Flip Over & 0.0 & 52.2 & 0.0 & 78.3 & 100 & 46.1 \\
      \midrule
      \multirow{4}{*}{Screwdriver} & \multirow{2}{*}{Long} & Spin
      Vert & 20.9 & 72.1 & 53.5 & 74.4 & 58.1 & 55.8 \\
      &  & Spin Horiz & 48.4 & 86.7 & 83.3 & 3.3 & 86.7 & 61.7 \\
      \cmidrule{2-9}
      & \multirow{2}{*}{Short} & Spin Vert & 15.4 & 61.5 & 12.8 & 100
      & 0.0 & 37.9 \\
      &  & Spin Horiz & 100 & 100 & 100 & 77.8 & 0.0 & 75.6 \\
      \bottomrule
    \end{tabular}%
  }
\end{table*}

\subsection{\textit{DexToolBench} Details}
\label{app:dex_tool_bench}

\begin{figure}[h!]
  \centering
  \includegraphics[width=0.5\textwidth]{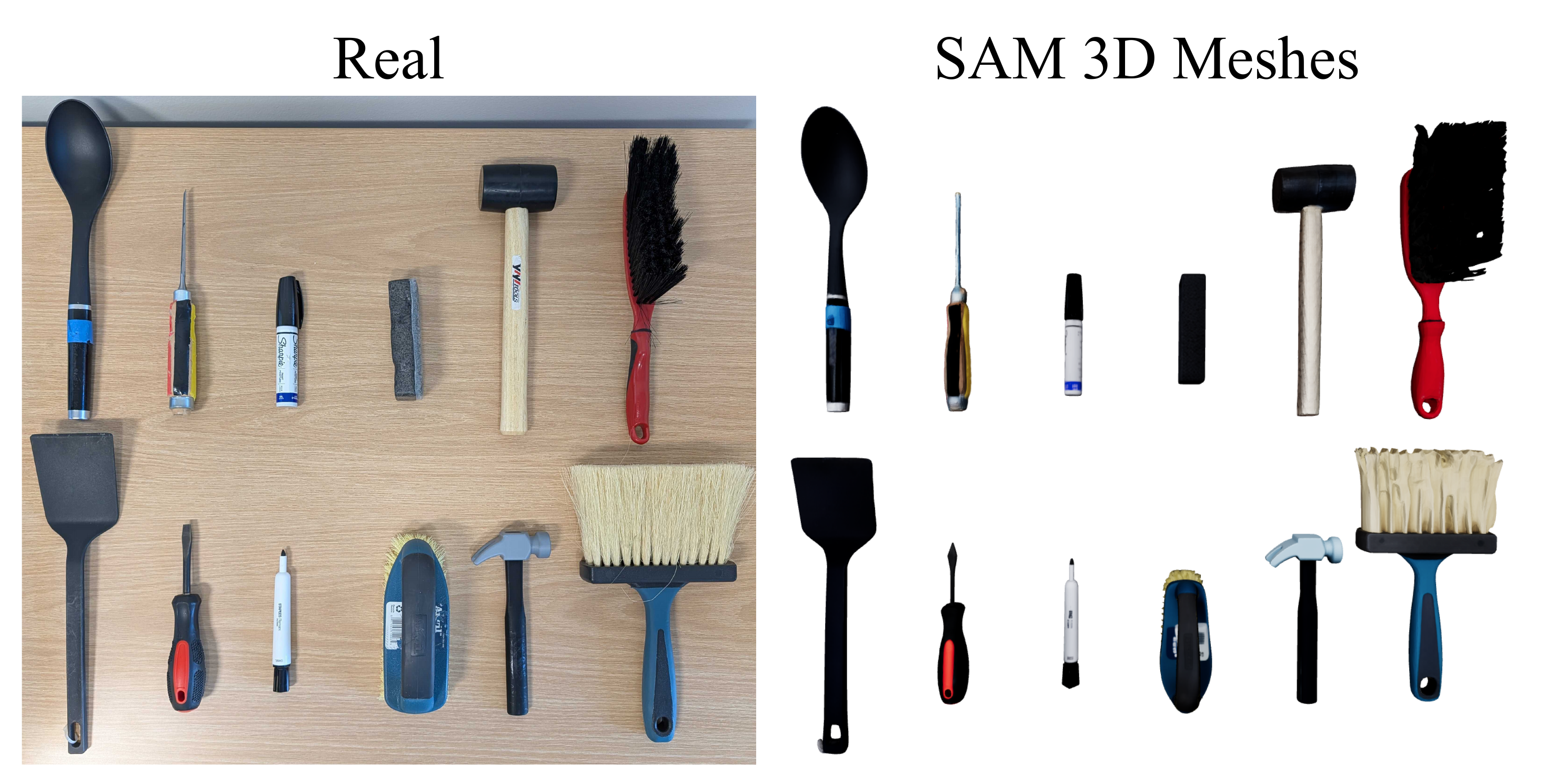}
  \captionsetup{width=0.5\textwidth}
  \caption{\textbf{\textit{DexToolBench} Objects.} (Left): Real-world
    objects. (Right): SAM 3D~\cite{chen2025sam} generated meshes of
  these objects.}
  \label{fig:real_sim_objects}
\end{figure}

Fig.~\ref{fig:real_world_task_frames} visualizes tasks from
\textit{DexToolBench}. Fig.~\ref{fig:real_sim_objects} visualizes the
real-world objects and SAM 3D~\cite{chen2025sam} generated objects in
\textit{DexToolBench}.

\textbf{Objects.} Here, we describe the 12 object instances across
the 6 object categories:
\begin{itemize}
  \item \textbf{Hammer}:
    \begin{itemize}
      \item \textit{Claw Hammer}: A 3D printed hammer consisting of a
        thin black handle and a gray head.
      \item \textit{Mallet Hammer}: A rubber mallet consisting of a
        wooden handle and a heavy, cylindrical black head.
    \end{itemize}
  \item \textbf{Marker}:
    \begin{itemize}
      \item \textit{Sharpie Marker}: A standard Sharpie permanent
        marker with a black cap.
      \item \textit{Staples Marker}: A dry-erase Staples marker with
        a white barrel. It is slightly thinner than the Sharpie.
    \end{itemize}
  \item \textbf{Eraser}:
    \begin{itemize}
      \item \textit{Handle Eraser}: An eraser with a thin handle and
        yellow bristles.
      \item \textit{Flat Eraser}: An Expo eraser consisting of a rectangular foam block without a handle.
    \end{itemize}
  \item \textbf{Brush}:
    \begin{itemize}
      \item \textit{Blue Brush}: A blue brush with a thick black
        handle and yellow bristles along the same direction as the handle.
      \item \textit{Red Brush}: A red brush with a long black handle
        and black bristles at a 90$^\circ$ angle with respect to the
        handle. It is lighter than the Blue brush.
    \end{itemize}
  \item \textbf{Spatula}:
    \begin{itemize}
      \item \textit{Spoon Spatula}: A spoonula with a long,
        cylindrical black handle and a shallow oval spoon.
      \item \textit{Flat Spatula}: A black spatula with a flat
        rectangular blade and thin, rectangular handle.
    \end{itemize}
  \item \textbf{Screwdriver}:
    \begin{itemize}
      \item \textit{Long Screwdriver}: A screwdriver with a long
        shaft and wooden handle. Alternating yellow, black, and red
        pieces of tape are added to the handle to reduce rotational symmetry.
      \item \textit{Short Screwdriver}: A screwdriver with a short
        shaft and a bulbous handle with a red and black pattern.
    \end{itemize}
\end{itemize}

\textbf{Tasks.} Here, we describe the 12 tool-use task trajectories
instances across the 6 object categories:

\begin{itemize}
  \item \textbf{Hammer}:
    \begin{itemize}
      \item \textit{Swing Down}: Grasp the hammer from flat on the
        table, rotate it by 90$^\circ$ into a striking configuration,
        swing down onto a nail 3 times.
      \item \textit{Swing Side}: Grasp the hammer from flat on the
        table, rotate it by 90$^\circ$ into a striking configuration,
        swing sideways onto a nail 3 times.
    \end{itemize}

  \item \textbf{Marker}:
    \begin{itemize}
      \item \textit{Draw Smile}: Grasp the marker from flat on the
        table, move to the whiteboard, draw two dots and a smile.
      \item \textit{Write C}: Grasp the marker from flat on the
        table, move to the whiteboard (different location from smile), draw a C.
    \end{itemize}

  \item \textbf{Eraser}:
    \begin{itemize}
      \item \textit{Wipe Smile}: Grasp the eraser from flat on the
        table, move to the whiteboard, erase the smile.
      \item \textit{Wipe C}: Grasp the eraser from flat on the table,
        move to the whiteboard (different location from smile), erase the C.
    \end{itemize}

  \item \textbf{Brush}:
    \begin{itemize}
      \item \textit{Sweep Forward}: Grasp the brush from flat on the
        table, rotate it by 90$^\circ$ into a sweeping configuration,
        sweep forward 3 times to sweep paper balls into trash.
      \item \textit{Sweep Right}: Grasp the brush from flat on the
        table, rotate it by 90$^\circ$ into a sweeping configuration,
        sweep rightwards 3 times to sweep paper balls into trash.
    \end{itemize}

  \item \textbf{Spatula}:
    \begin{itemize}
      \item \textit{Serve Plate}: Grasp the spatula from flat on the
        plate. Move to the other bowl/plate, perform a scooping
        motion, perform a serving motion onto the original plate.
      \item \textit{Flip Over}: Grasp the spatula upside-down with
        the spatula head on the pan/bowl. Rotate it by 180$^\circ$,
        perform scooping motion, perform flipping motion.
    \end{itemize}

  \item \textbf{Screwdriver}:
    \begin{itemize}
      \item \textit{Spin Vertical}: Grasp the screwdriver from flat
        on the table, rotate it by 90$^\circ$ into a vertical
        configuration, spin it by 360$^\circ$ along long axis.
      \item \textit{Spin Horizontal}: Grasp the screwdriver from flat
        on the table, keep it at a horizontal configuration, spin it
        by 360$^\circ$ along long axis.
    \end{itemize}
\end{itemize}

\textbf{Processed Data.} Each task is defined by an RGB-D human
video, captured via a ZED 1 stereo camera. We process each video with
our perception pipeline, described in detail in
Appendix~\ref{app:human_video_processing}. The processed data includes:
\begin{itemize}
  \item Raw RGB-D video
  \item Object segmentation masks
  \item Metric-scale object meshes
  \item 6D object pose trajectories
\end{itemize}

Additionally, we provide:
\begin{itemize}
  \item Visualization scripts to visualize the data in 3D space using
    Viser~\cite{yi2025viser}.
  \item Simulation scripts to evaluate policies on these tasks. This
    will support multiple robot arms and hands.
  \item Links to purchase the real-world objects
\end{itemize}

\textbf{\textit{Task Progress}.} For each task, we evaluate
\textit{Task Progress}, defined as the percentage of object pose
waypoints in the demonstrated trajectory that the robot successfully
reached. We consider a goal pose to be reached if the distance
between the object pose and goal pose $d(\bm{o}_t,\bm{g})$ is below
our defined success tolerance $\epsilon=2cm$.

This evaluation is performed \textit{closed-loop}: the goal advances
to the next waypoint only after the current waypoint is reached. This
contrasts with \textit{open-loop} playback, where goals advance at a
fixed frequency regardless of the object's state. Our closed-loop
protocol decouples execution speed from spatial accuracy, allowing
the policy to utilize retry behaviors and execute as fast or slow as
it needs to without being penalized for failing to match the exact
timing profile of the demonstration.

\begin{figure*}[t!]
  \centering
  \includegraphics[width=0.95\textwidth]{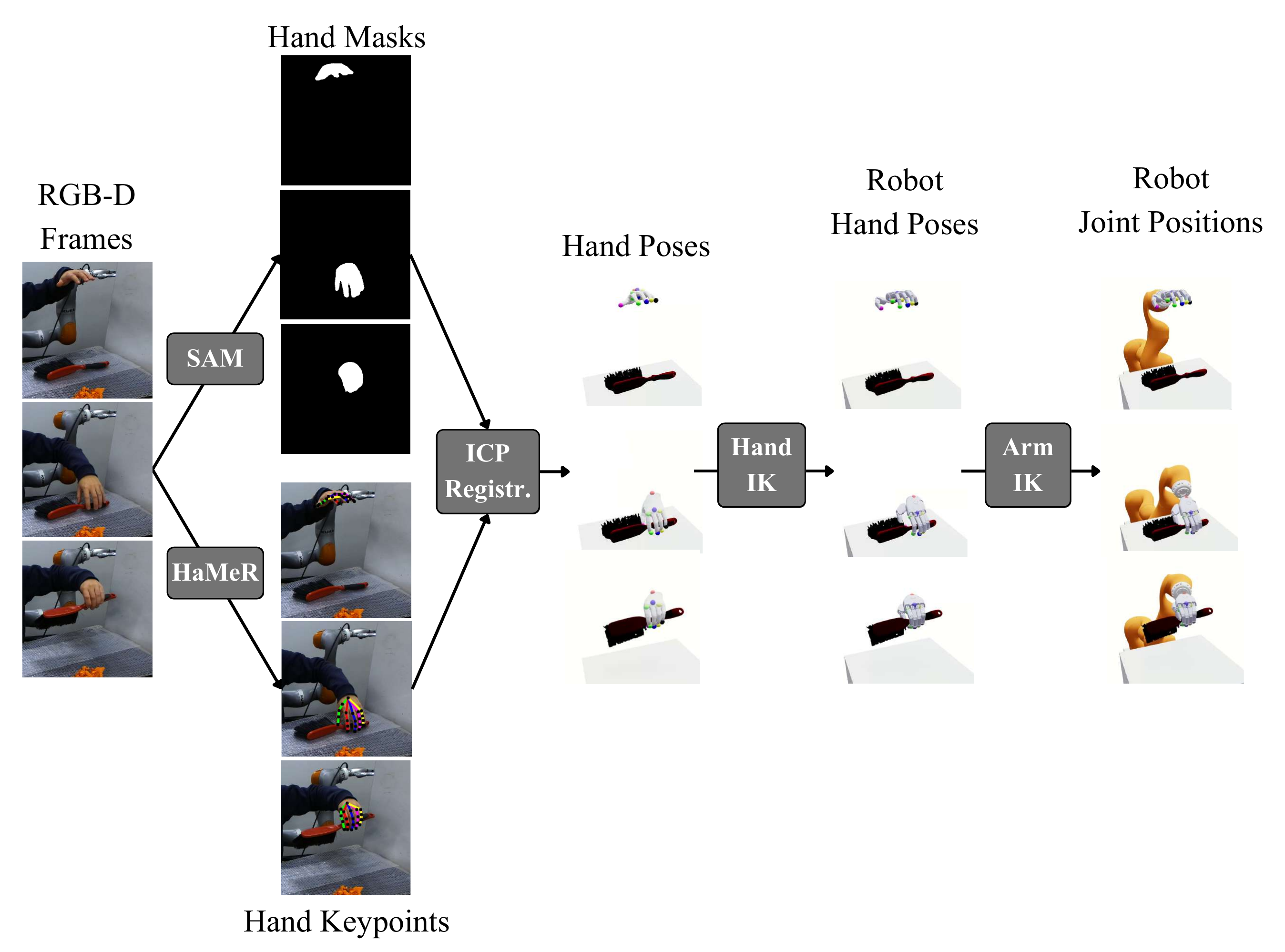}
  \captionsetup{width=\textwidth}
  \caption{\textbf{Kinematic Retargeting Pipeline.} From the RGB-D
    human video, we use SAM 2~\cite{ravi2024sam2} for hand masks and
    HaMeR~\cite{pavlakos2024reconstructing} for hand pose prediction.
    Next, we use ICP registration to align the hand pose prediction
    with the segmented hand point cloud to obtain accurate 3D hand
    poses. Lastly, we perform IK-based retargeting of the arm and hand
  to match the human wrist pose and fingertip positions.}
  \label{fig:kinematic_retargeting_visualization}
\end{figure*}

\subsection{Real-World Experiment Additional Analysis}
\label{app:real_world_analysis}

Table~\ref{tab:detailed_results} shows the per-rollout \textit{Task
Progress} of the results shown in
Fig.~\ref{fig:generalizationResults}.
Fig.~\ref{fig:visualized_policy_observations} visualizes the policy
observations at each timestep during real-world deployment.

\textbf{Perception Failures.} We observe that object pose estimation
is the most common failure mode of our system (43.7\% of failures).
Specifically, FoundationPose~\cite{wen2024foundationpose} struggles
under three conditions common in dexterous manipulation: (1) heavy
object occlusion, which is exacerbated when the robot hand
manipulates small objects (e.g., the small screwdriver); (2) visual
ambiguity due to rotational symmetry (e.g., the cylindrical body of
the marker); and (3) low visual contrast between the object and the
environment (e.g., dark bristles on a black surface). While our
policy is trained to be robust to significant pose noise, it cannot
recover from catastrophic tracking failures where the estimator loses
the object entirely. Future work could address this by incorporating
additional camera views or leveraging temporal consistency in 2D
tracking to improve state estimation stability.

\textbf{Manipulation Failures.} Manipulation failures primarily stem
from two sources: object drops (34.5\%) and failure to reach the goal
pose due to incomplete in-hand rotation (18.2\%). Object drops were
most common on heavy objects (e.g., the 331g \texttt{mallet hammer}
and the 325g \texttt{blue brush}), typically occurring during
reorientation or environment contact. Failure to reach the goal pose
due to incomplete in-hand rotation was most common on thin objects
(e.g., the $\sim$1cm thick \texttt{flat spatula}), as the policy
would repeatedly attempt in-hand rotation but fail to manipulate the
thin geometry. Grasp failures were rare (3.6\%) but occasionally
occurred with the marker due to its thin shape. In these instances,
the marker's cylindrical shape often caused it to roll off of the
table, so the policy was unable to recover.

\subsection{Kinematic Retargeting Baseline Details}
\label{app:kinematic_retargeting}

Fig.~\ref{fig:kinematic_retargeting_visualization} visualizes the
kinematic retargeting pipeline. Following
\citet{lum2025crossinghumanrobotembodimentgap}, we perform hand pose
estimation on each frame of the human video and then retarget these
hand poses to a dexterous robot. Specifically, we first use
HaMeR~\cite{pavlakos2024reconstructing} to predict initial hand
keypoints and mesh vertices from RGB images. Next, we refine this
prediction by using the corresponding depth image and a mask of the
hand (from SAM~2~\cite{ravi2024sam2}) to extract a 3D hand point
cloud, and then aligning the HaMeR mesh to this point cloud via
Iterative Closest Point (ICP) registration. Finally, we retarget the
robot to these poses in two stages: we first compute the arm
configuration using Damped Least Squares~\cite{buss2004introduction}
to place the palm, and subsequently optimize the hand finger joints
to reach the fingertip targets relative to the palm.

\subsection{Fixed-Grasp Baseline Details}
\label{app:fixed_grasp}

\begin{figure}[h!]
  \centering
  \includegraphics[width=0.5\textwidth]{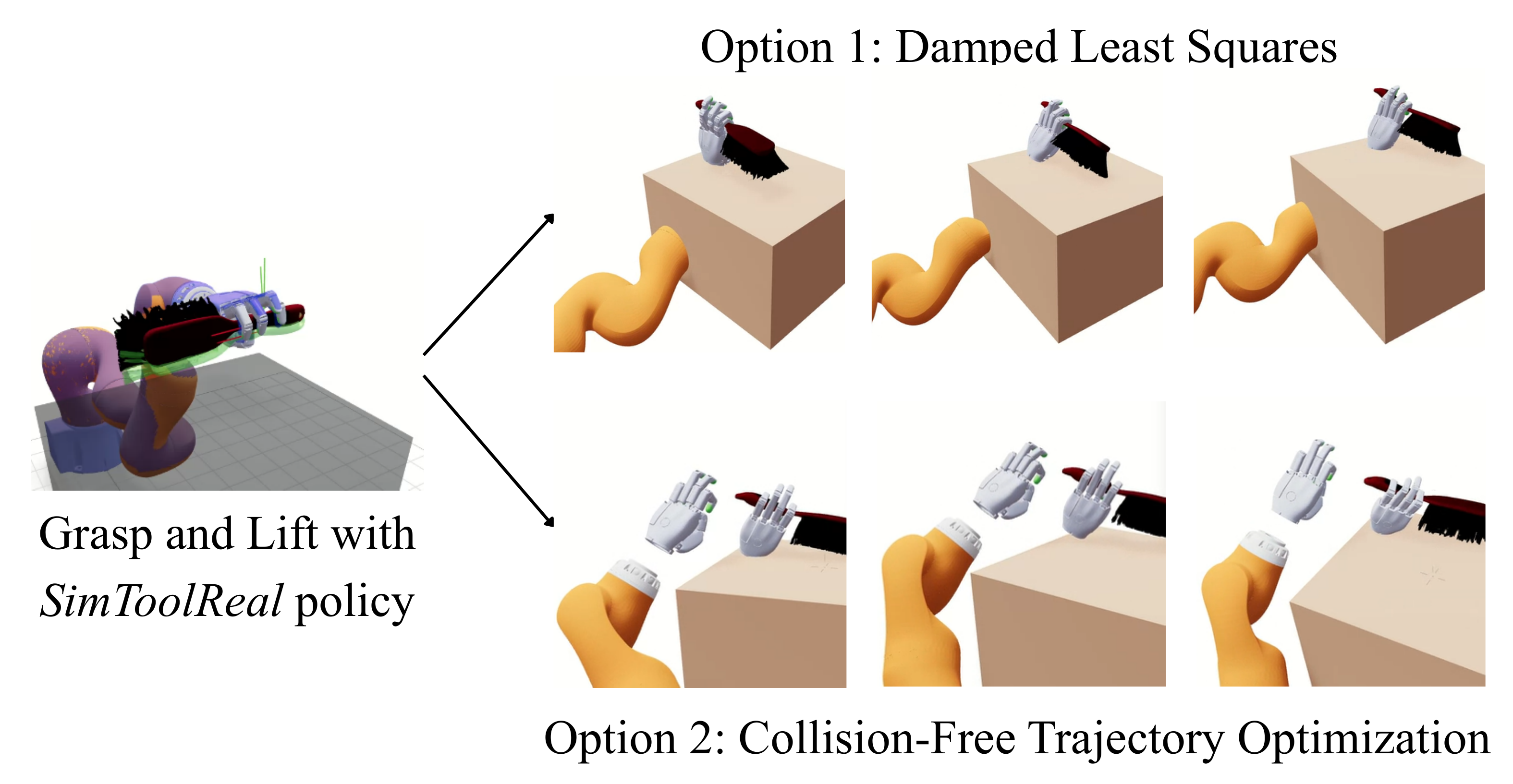}
  \captionsetup{width=0.5\textwidth}
  \caption{\textbf{Fixed Grasp Baselines.} We first use our
    \textit{SimToolReal} policy to grasp and lift the object. We then
    attempt to follow the trajectory using a fixed grasp. Option 1
    (Damped Least Squares): Tracks the target poses but causes severe
    collisions with the table. Option 2 (Collision-Free Trajectory
  Optimization): Avoids collisions but fails to reach the target poses.}
  \label{fig:fixed_grasp_visualization}
\end{figure}

To run this baseline, we first use our \textit{SimToolReal} policy to
grasp and lift the object to the initial goal pose. Once the first
goal pose is reached, we stop running the policy and begin planning a
fixed grasp trajectory to perform the object-trajectory following
task. For the hand, we maintain a fixed grasp by storing the current
hand joint position targets and continuing to apply these joint
position targets to the hand. For the arm, we plan a motion that
moves the object along the goal trajectory, assuming its relative
pose to the end-effector remains constant.

\textbf{Problem Formulation.} Specifically, we assume a rigid grasp
where the transformation $T_{EO}$ between the end-effector frame $E$
and the object frame $O$ remains constant throughout the trajectory.
Let $T_{BO}^{(i)}$ denote the $i$-th target object pose in the robot
base frame $B$ from a goal sequence of length $N$. The corresponding
target end-effector pose $T_{BE}^{(i)}$ is derived as:
\begin{equation}
  T_{BE}^{(i)} = T_{BO}^{(i)} \left(T_{EO}\right)^{-1}
\end{equation}
We test two methods for computing the arm joint trajectory
$\bm{q}_{1:N}$ that reach these target end-effector poses.

\begin{itemize}[leftmargin=*]
  \item \textbf{Damped Least Squares.}
    For each waypoint, we compute the arm Jacobian $J(\bm{q})$ and
    the pose error vector $\bm{e}$ (concatenating translation and
    axis-angle rotation error). We compute the iterative joint update
    $\Delta \bm{q}$ using the Damped Least Squares
    method~\cite{buss2004introduction}:
    \begin{equation}
      \Delta \bm{q} = J^T (JJ^T + \lambda^2 I)^{-1} \bm{e}
    \end{equation}
    where the damping factor $\lambda$ smooths the motion and
    prevents discontinuities. While this method is computationally
    efficient (solving in seconds), it is strictly local and does not
    account for environmental obstacles (e.g., the table surface or
    whiteboard), often leading to collisions.

  \item \textbf{Collision-Free Trajectory Optimization.}
    To address environment collisions, we utilize
    PyRoki~\cite{kim2025pyroki}, which uses a JAX-based
    Levenberg-Marquardt solver for trajectory optimization. We extend
    its standard trajectory optimization formulation to optimize a
    trajectory with $N$ waypoints instead of 1. This method incurs a
    higher computation time ($\sim$40--60 seconds per trajectory) and
    requires a time-consuming process of modeling the robot as a set
    of collision spheres.
    \begin{enumerate}
      \item \textit{Initialization:} We warm-start the optimization
        by sequentially solving collision-free Inverse Kinematics
        (IK) for each waypoint, seeding each step with the solution
        from the previous waypoint to encourage temporal consistency.
      \item \textit{Optimization:} The solver minimizes a cost
        function composed of target pose error and smoothness (joint
        velocity) terms, subject to constraints: joint
        position/velocity limits, self-collision avoidance, and
        environment collision avoidance.
    \end{enumerate}
\end{itemize}

Fig.~\ref{fig:fixed_grasp_visualization} visualizes the trajectories
produced by both methods. We use the Collision-Free Trajectory
Optimization for all reported baseline experiments, as the simple
Damped Least Squares method frequently fails due to significant table
collisions during complex object rotations.

\subsection{Specialist Details}
\label{app:specialist}

\begin{figure*}[t!]
  \centering
  \includegraphics[width=0.41\textwidth]{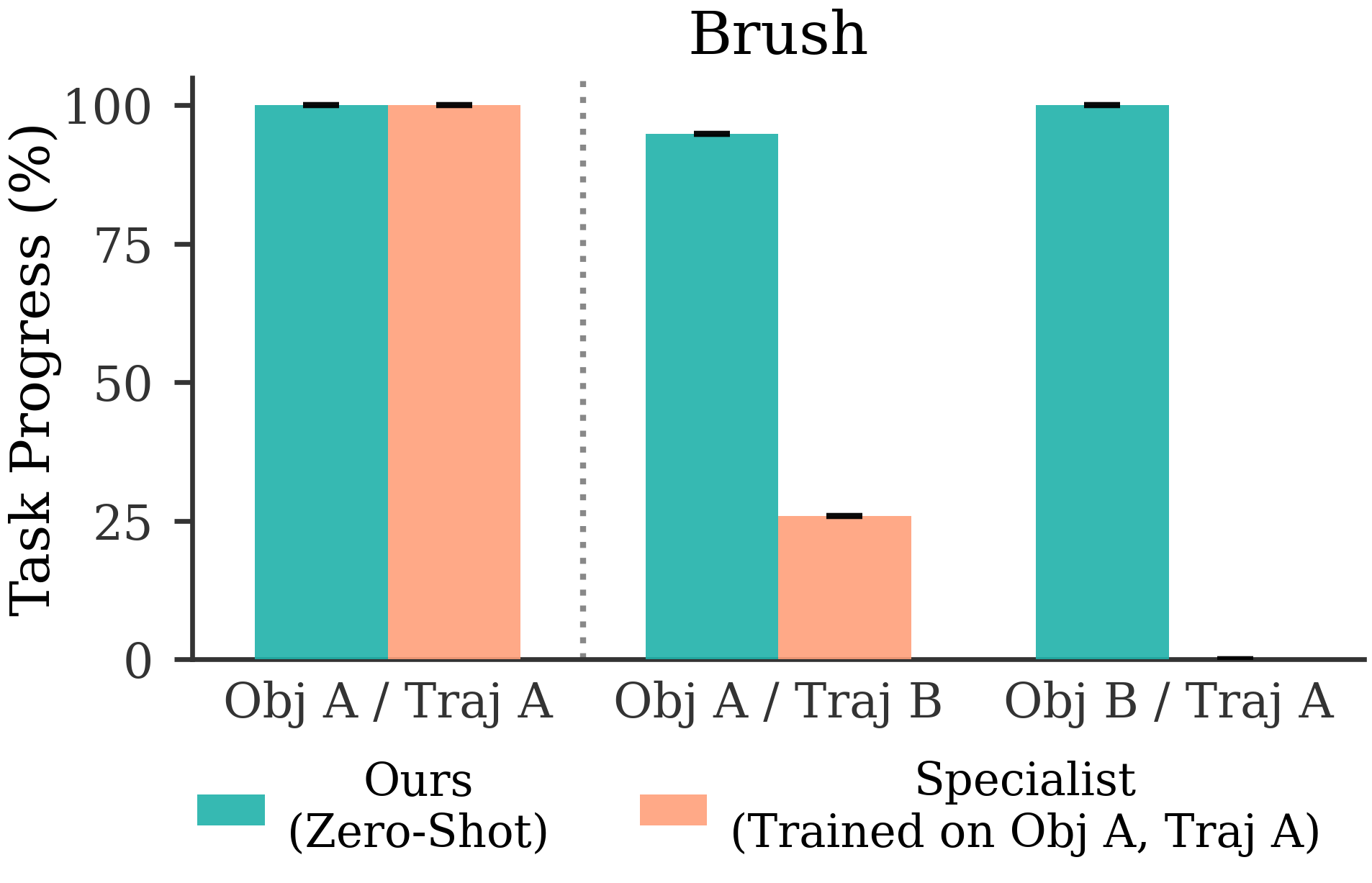}
  \includegraphics[width=0.41\textwidth]{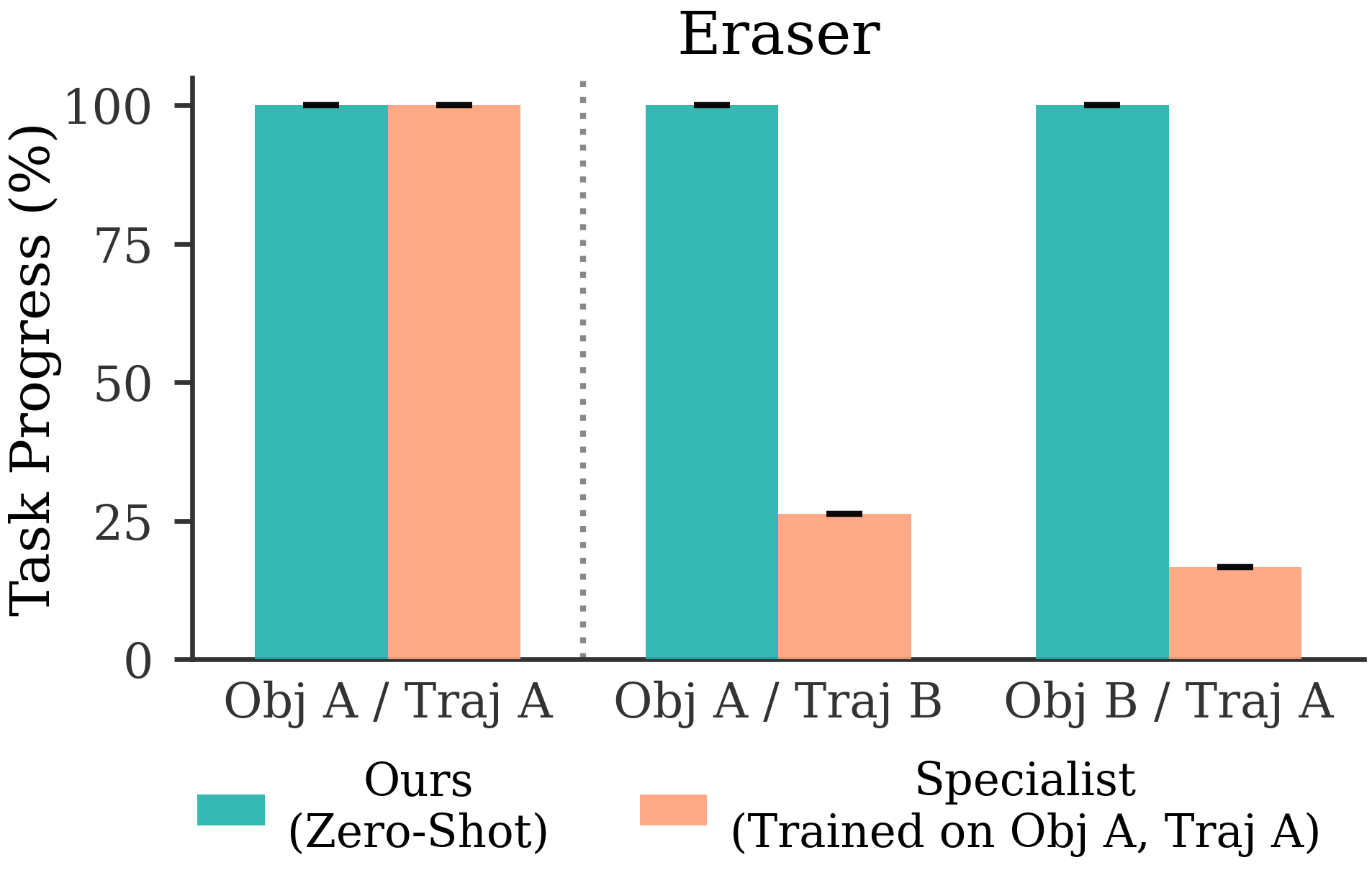}
  \includegraphics[width=0.41\textwidth]{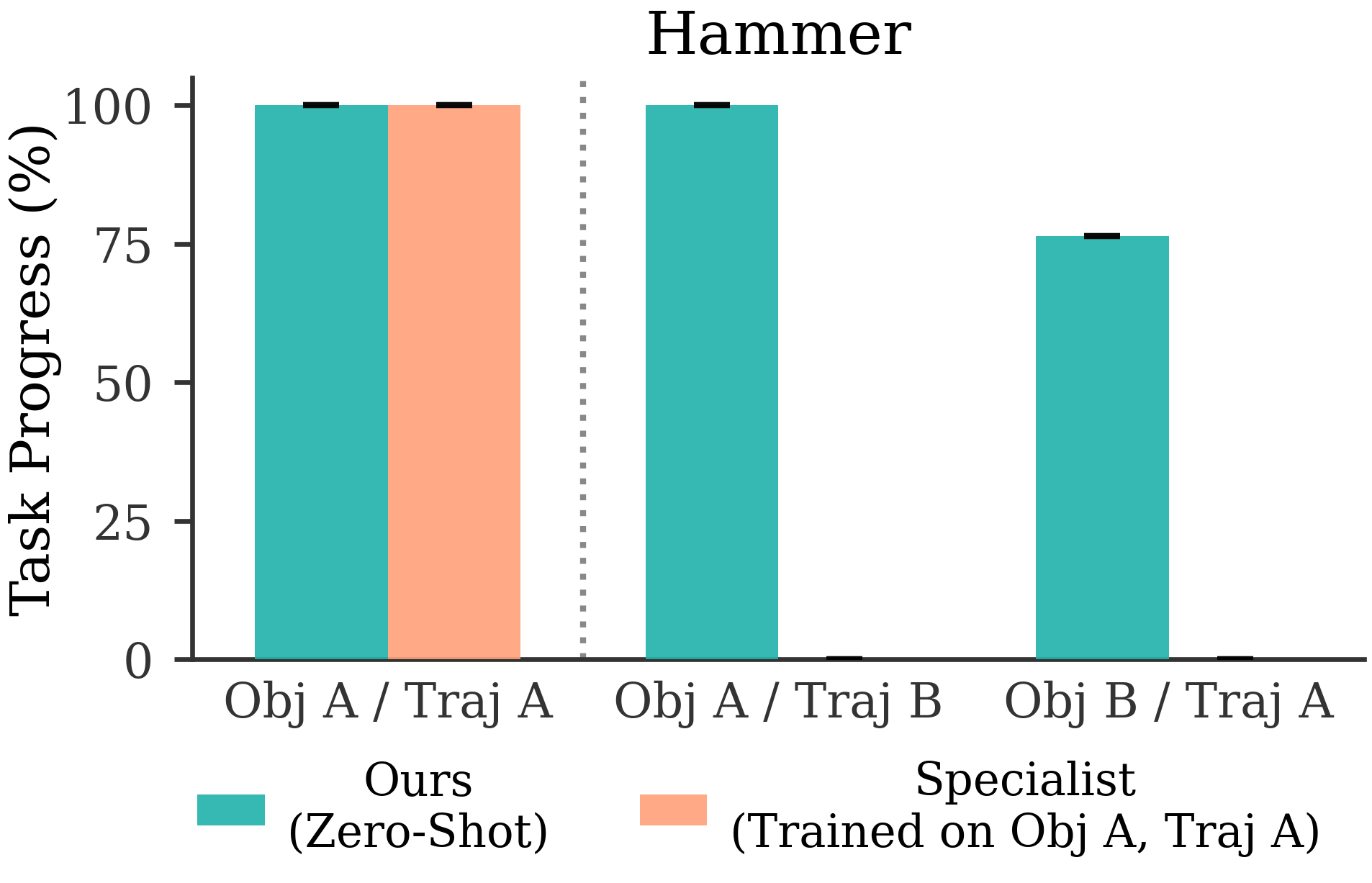}
  \includegraphics[width=0.41\textwidth]{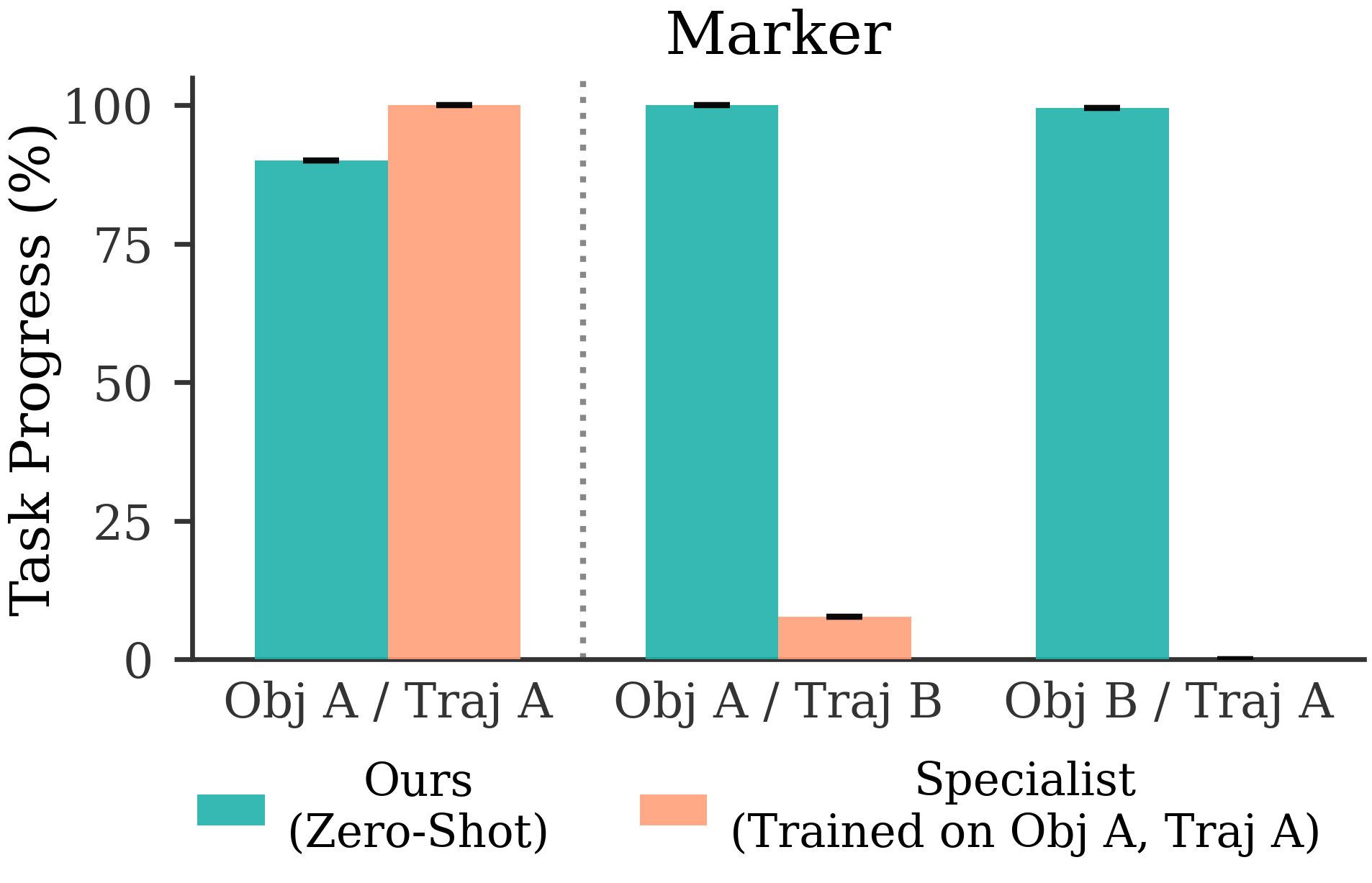}
  \includegraphics[width=0.41\textwidth]{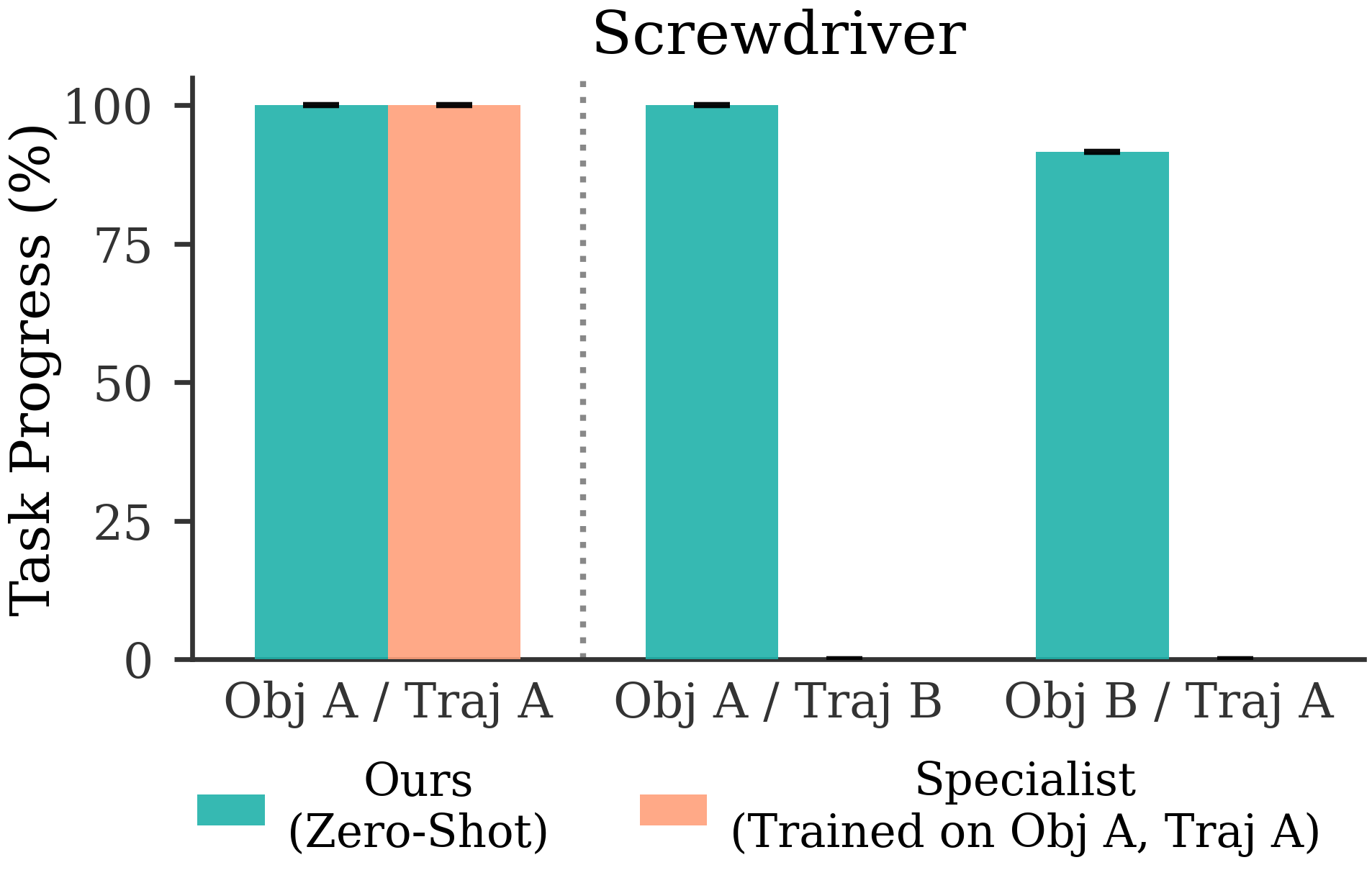}
  \includegraphics[width=0.41\textwidth]{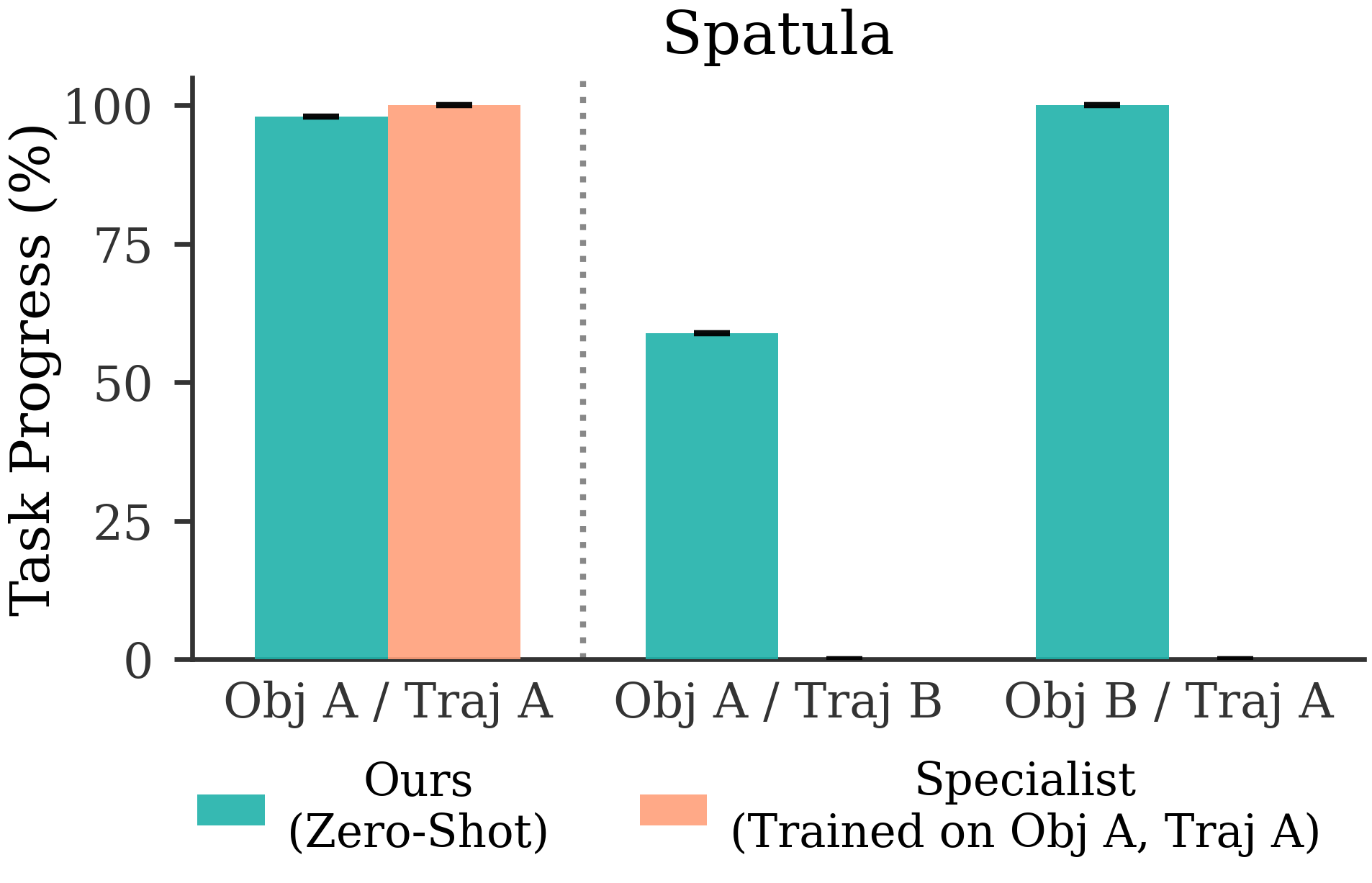}
  \captionsetup{width=\textwidth}
  \caption{\textbf{Detailed Comparison against Specialists.} We
    provide a breakdown of the results in
    Fig.~\ref{fig:specialistComparison}. Each plot compares
    \textit{SimToolReal} to a specialist policy trained on a single
  object and trajectory (Obj A / Traj A) for a single object category.}
  \label{fig:specialist_all_reuslts}
\end{figure*}

Fig.~\ref{fig:specialist_all_reuslts} presents a granular breakdown
of the aggregate results shown in
Fig.~\ref{fig:specialistComparison}. While the main text reports the
average performance across all categories, this figure details the
specific generalization capabilities of the specialist policy trained
for each of the six tool-use categories individually (e.g., Hammer,
Spatula, Brush). Each specialist policy uses the same architecture,
observation space, and action space as used by \textit{SimToolReal}.
It uses the same reward function, termination criteria, and episode
initialization settings. The only changes are (i) \textit{Fixed
Object Geometry}: instead of training on diverse, procedurally
generated primitive objects, the specialist is trained exclusively on
a single object instance (Obj A), and (ii) \textit{Fixed Task
Trajectory}: instead of training on randomly sampled goal poses, the
specialist is trained on a single, fixed sequence of goal poses (Traj
A) extracted from a specific human demonstration.

Table~\ref{tab:specialist_definitions} details the specific object
instances and task trajectories that define Obj A, Traj A, Obj B, and
Traj B for each tool category.

\begin{table}[tb]
  \centering
  \caption{\textbf{Specialist Objects and Trajectories.} Objects and
  trajectories used for specialist policy training and evaluation.}
  \resizebox{\columnwidth}{!}{
    \begin{tabular}{@{}lllll@{}}
      \toprule
      \textbf{Category} & \textbf{Obj A} & \textbf{Traj A} &
      \textbf{Obj B} & \textbf{Traj B} \\
      \midrule
      \multirow{2}{*}{Brush}       & Red         & Sweep      & Blue
      & Sweep \\
      & Brush       & Forward    & Brush       & Right \\
      \midrule
      \multirow{2}{*}{Eraser}      & Flat        & Wipe       &
      Handle      & Wipe \\
      & Eraser      & Smile      & Eraser      & C \\
      \midrule
      \multirow{2}{*}{Hammer}      & Mallet      & Swing      & Claw
      & Swing \\
      & Hammer      & Down       & Hammer      & Side \\
      \midrule
      \multirow{2}{*}{Marker}      & Staples     & Draw       &
      Sharpie     & Write \\
      & Marker      & Smile      & Marker      & C \\
      \midrule
      \multirow{2}{*}{Screwdriver} & Long        & Spin       & Short
      & Spin \\
      & Screwdriver & Vertical   & Screwdriver & Horizontal \\
      \midrule
      \multirow{2}{*}{Spatula}     & Flat        & Serve      & Spoon
      & Flip \\
      & Spatula     & Plate      & Spatula     & Over \\
      \bottomrule
    \end{tabular}%
  }
  \label{tab:specialist_definitions}
\end{table}

\end{document}